\PassOptionsToPackage{sort&compress}{natbib}

\documentclass[preprint,3p,12pt]{elsarticle}

\usepackage{amssymb}
\usepackage{tabularray}
\usepackage{placeins}

\usepackage{booktabs}

\usepackage{xcolor}
\usepackage{graphicx}
\usepackage{subfigure}
\journal{Additive Manufacturing}

\begin{document}

\begin{frontmatter}



\title{Inexpensive High Fidelity Melt Pool Models in Additive Manufacturing Using Generative Deep Diffusion }
%Super-Resolution of Laser Powder Bed Fusion Melt Pool Dynamics using Generative Diffusion Models
%alt title: High Fidelity Inexpensive Model of Meltpool Obtained Using Deep Diffusion Models

\author[inst1]{Francis Ogoke \corref{contrib}}
\affiliation[inst1]{organization={Department of Mechanical Engineering, Carnegie Mellon University},%Department and Organization
            %addressline={5000 Forbes Avenue}, 
            city={Pittsburgh},
            postcode={15213}, 
            state={PA},
            country={USA}}

\author[inst4]{Quanliang Liu \corref{contrib}}
\author[inst1]{Olabode Ajenifujah}
\author[inst1]{Alexander Myers}
\author[inst1]{Guadalupe Quirarte}
\author[inst1]{Jack Beuth}
\author[inst1]{Jonathan Malen}

\author[inst1,inst2,inst3]{Amir Barati Farimani\corref{corauthor}}
\affiliation[inst2]{organization={Department of Chemical Engineering, Carnegie Mellon University},%Department and Organization
            %addressline={5000 Forbes Avenue}, 
            city={Pittsburgh},
            postcode={15213}, 
            state={PA},
            country={USA}}
\affiliation[inst3]{organization={Machine Learning Department, Carnegie Mellon University},%Department and Organization
            %addressline={5000 Forbes Avenue}, 
            city={Pittsburgh},
            postcode={15213}, 
            state={PA},
            country={USA}}
\affiliation[inst4]{organization={Department of Materials Science and Engineering, Carnegie Mellon University},%Department and Organization
            %addressline={5000 Forbes Avenue}, 
            city={Pittsburgh},
            postcode={15213}, 
            state={PA},
            country={USA}}
\begin{abstract}
%%To-do: rephrase to avoid conflict with APS abstract
 Defects in laser powder bed fusion (L-PBF) parts often result from the  meso-scale dynamics of the molten alloy near the laser, known as the melt pool. For instance, the melt pool can directly contribute to the formation of undesirable porosity, residual stress, and surface roughness in the final part. Experimental in-situ monitoring of the three-dimensional melt pool physical fields is challenging, due to the short length and time scales involved in the process. Multi-physics simulation methods can describe the three-dimensional dynamics of the melt pool, but are computationally expensive at the mesh refinement required for accurate predictions of complex effects, such as the formation of keyhole porosity. Therefore, in this work, we develop a generative deep learning model based on the probabilistic diffusion framework to map low-fidelity, coarse-grained simulation information to the high-fidelity counterpart. By doing so, we bypass the computational expense of conducting multiple high-fidelity simulations for analysis by instead upscaling lightweight coarse mesh simulations. Specifically, we implement a 2-D diffusion model to spatially upscale cross-sections of the coarsely simulated melt pool to their high-fidelity equivalent. We demonstrate the preservation of key metrics of the melting process between the ground truth simulation data and the diffusion model output, such as the temperature field, the melt pool dimensions and the variability of the keyhole vapor cavity. Specifically, we predict the melt pool depth within 3 $\mu m$ based on low-fidelity input data  4$\times$ coarser than the high-fidelity simulations, reducing analysis time by two orders of magnitude.

\end{abstract}


% \begin{highlights}
% \end{highlights}


\end{frontmatter}


\section{Introduction}
\label{sec:sample1}
Laser Powder Bed Fusion (L-PBF) is an Additive Manufacturing (AM) process that enables the construction of parts with complex geometries. During L-PBF, a heat source melts and fuses successive layers of microscopic metallic powder. This production paradigm enables the fabrication and rapid prototyping of parts difficult to manufacture with conventional machining methods \cite{king2015laser, sing2020laser}. However, widespread use of L-PBF techniques for production is limited by the tendency for defects to form during the process. The random accumulation of these defects induces a variability issue that affects the adoption of L-PBF for high-precision applications \cite{khairallah2016laser, gordon2020defect, mukherjee2018mitigation}. These defects are often dependent on the behavior of the mesoscale dynamics of the interactions of the laser heat source with the substrate \cite{panwisawas2017mesoscale, khairallah2020controlling, ogoke2021thermal}. During the melting process, the localized melting dynamics induce a recoil pressure on the free molten surface, causing a depression known as the keyhole cavity \cite{cunningham2019keyhole, zhao2017real}. Simultaneously, the laser heat input creates a localized melting zone known as the melt pool \cite{ur2021full}. In scenarios with excess heat input, the vapor cavity will become unstable and periodically collapse, trapping gas bubbles in the melt pool that solidify into pores \cite{shrestha2019numerical, huang2022keyhole}. However, in scenarios with low heat input, the melt pool will fail to overlap effectively, creating unmelted void structures within the final part. Therefore, it is necessary to investigate and optimize the operating conditions and processing parameters before printing to avoid the formation of excess porosity. Analytical models have been proposed to describe the dynamics of the melt pool, and are successful in low energy density regimes, but these assumptions fail to be upheld in cases with high energy density where the fundamental physics of the problem differ \cite{rosenthal1941mathematical, eagar1983temperature, promoppatum2017comprehensive, akbari2022meltpoolnet}. While process map based correlations are effective for establishing regions for fully dense printing, the part geometry can alter the properties of the powder bed from the nominal scenario \cite{le2020discontinuity, schmidt2022support, grunewald2023support}. For instance, Han et al. observe defect formation in overhang structures, caused by the lack of support structures locally lowering the thermal conductivity of the part \cite{han2018manufacturability}. Therefore, high-fidelity simulations of the melt pool dynamics are often required to evaluate the defect formation potential of a combination of processing parameters. 

Several methods for high-fidelity simulations of the melt pool dynamics experienced during L-PBF have been proposed \cite{cheng2019computational, ninpetch2023multiphysics, ahsan2022global}.
% Similarly, machine learning enabled surrogate models have also been proposed to predict the dynamics of the process given the processing parameters as input, but these models may not accurately predict behavior in cases where the processing parameters are off-nominal due to differences in machinery, or systematic deviations induced by the physics of the process. 
 For instance, Cheng et al. use the CFD package FLOW-3D to demonstrate the influence of Marangoni convection on the properties of the melt flow induced by temperature dependent surface tension  \cite{cheng2019computational}. In a similar study, Ninpetch et al. implement a FLOW-3D model to investigate the influence of the thickness of the powder layer on the dimensions of the underlying melt pool \cite{ninpetch2023multiphysics}. However, these simulation tools are often time-consuming to implement at a sufficiently small mesh element size for parameter study, large-scale designs of experiments, and path-planning for complex geometries.

 The computational expense associated with high-fidelity simulations have motivated the use of deep learning tools for both global prediction and surrogate modeling in physical applications \cite{yang2019predictive, ogoke2021graph, wen2022u}. These advances have also been applied to accelerate simulations of the meso-scale laser material interactions, due to the runtime required to resolve the complicated multi-scale phenomena that emerges during the melting process \cite{hemmasian2023surrogate, strayer2022accelerating}. We seek to extend these methods by providing a framework for inference where low-fidelity simulations of the melt pool behavior can be correlated to high fidelity simulations for rapid inference across the L-PBF parameter space. This technique is commonly referred to as super-resolution in the computer vision and imaging fields.

Super-resolution has been applied for a variety of tasks in image processing \cite{farsiu2004fast, wang2020deep, dong2014learning}, and has been accelerated with the advent of deep learning. In a typical approach, this method attempts to predict the high resolution version of a image, given a down-scaled version, preserving as many of the high-resolution image details present in the target image as possible. The success of this approach enables rapid restoration of degraded image files, and efficient image compression \cite{cheng2018performance, cao2020lossless}.  The demonstrated ability to reduce the memory required for processing data has also led to the emergence of super-resolution and multi-fidelity tools for accelerating simulations in fluid mechanics \cite{deng2019super, liu2020deep, li2022using, fathi2020super}. For instance, Deng et al. developed super-resolution reconstruction methods to recreate the complex wake flow behind cylinders using a Generative Adversarial Network (GAN) based framework \cite{deng2019super}. While generative models are able to reconstruct high-frequency small scale details by learning the distribution of high-resolution images conditioned on their low-resolution inputs, they are difficult to reliably train and potentially unstable during the inference process \cite{li2021tackling, bang2021mggan, liu2019spectral}. Here, we make use of diffusion models, which iteratively remove noise from a randomly sampled vector to generate a synthetic sample \cite{ho2020denoising}. During training, the diffusion model learns the empirical data distribution with a Markovian random process. In related applications, diffusion models have also been applied to image super-resolution tasks in computer vision \cite{li2022srdiff, saharia2022image}. Others have recently extended diffusion models for super-resolution in the physics domain \cite{shu2023physics, jadhav2023stressd}. We seek to apply diffusion based super-resolution in approaches where applying PDE-derived physics based loss functions may be intractable, e.g.,  in complex multiphysics scenarios. To do so, we combine a multi-scale feature extraction CNN model with the diffusion model paradigm to first map the low fidelity input data to an approximation of the high fidelity cross-section, which is then resolved by the diffusion model to reconstruct stochastic details. Therefore, we are able to capture the intrinsic variability of the melt pool vapor interactions, such as the rapid keyhole oscillations that occur in high energy density conditions. By doing so, our model enables the rapid evaluation of new process conditions, reducing the time required to interpolate within and characterize the parameter space.

\section{Methodology}
\label{sec:sample2}

\begin{figure}[htbp!]
    \centering
    \includegraphics[width=\textwidth]{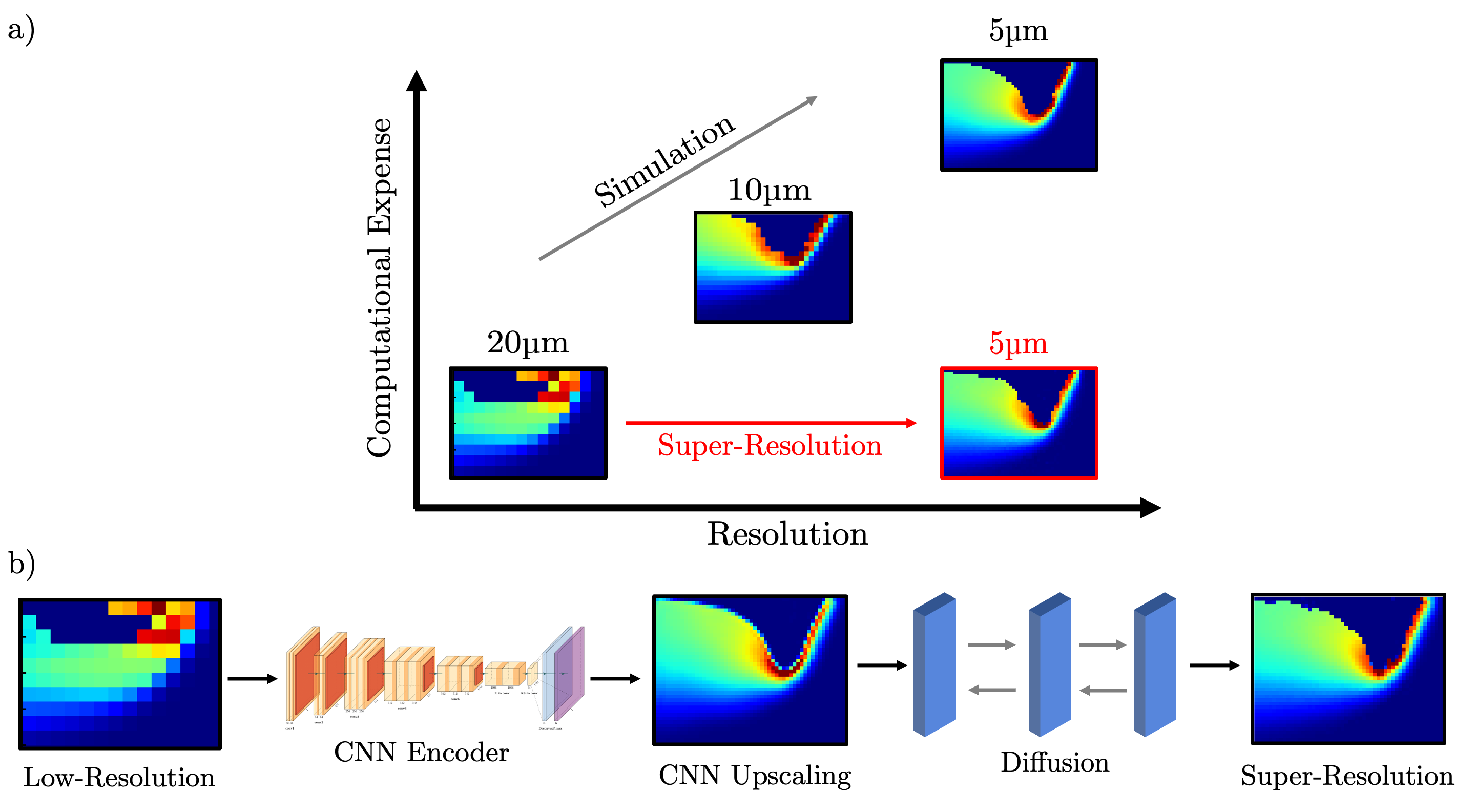}
    \caption{a) The goal of this model is to create a deep-learning method for upscaling simulations of laser powder bed fusion, reducing the computational time required to obtain high fidelity data given a dataset of coarsely resolved simulations. b) To do so, the physical fields of the low fidelity simulation first are passed through a ResNet based upscaling model, followed by a further refining process carried out by a conditional diffusion model.}
    \label{fig:pipeline}
\end{figure}

\begin{figure}[htbp!]
    \centering
    \includegraphics[width=\textwidth]{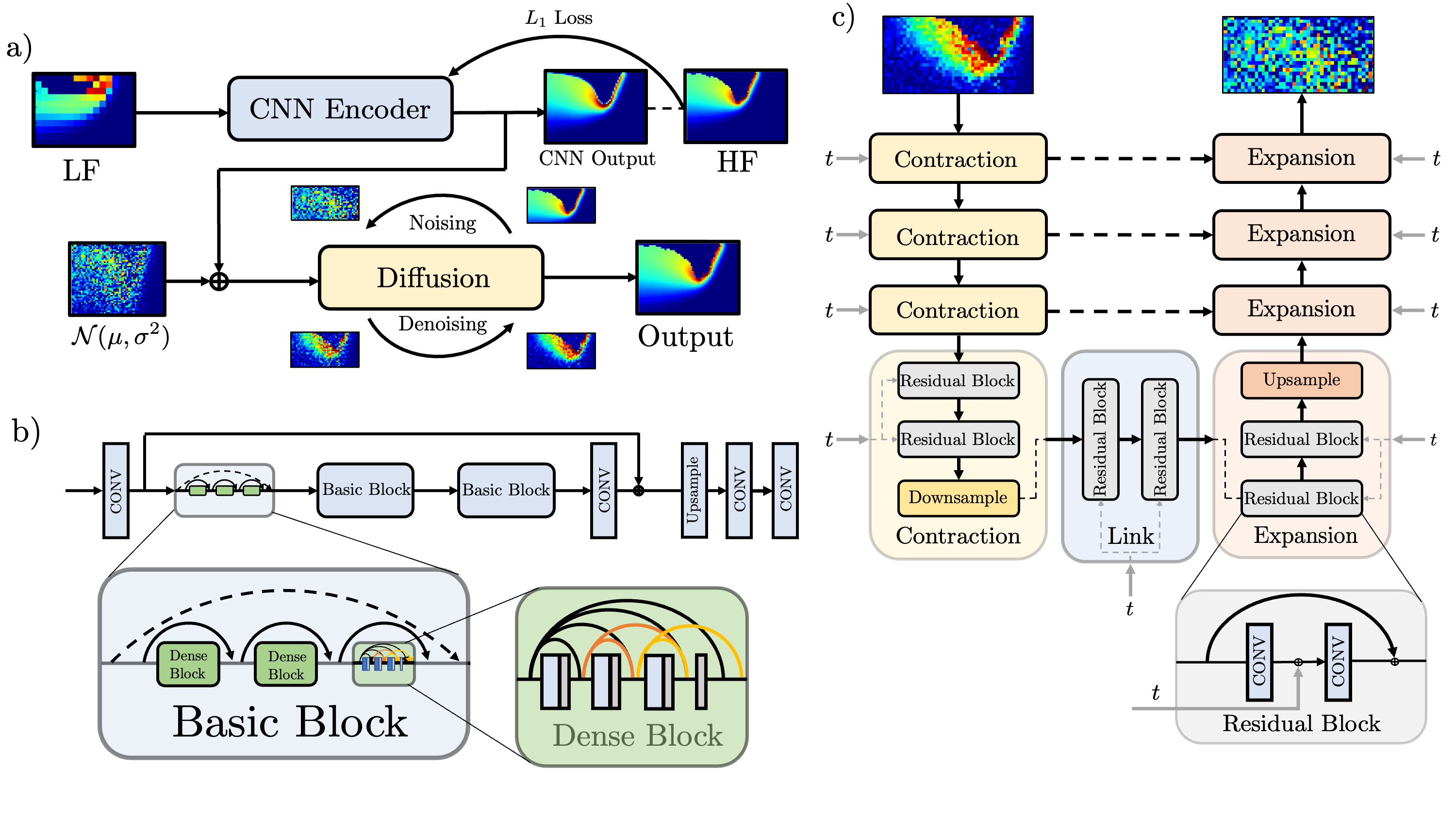}
    \caption{
    a) The overall model architecture. The low-fidelity cross-section is passed to a CNN model, which attempts to predict the average high-fidelity (HF) image from the low-fidelity (LF) input. This model is trained to minimize the $L_1$ loss between the HF target image and the model output. Next, a noise vector is sampled from the standard normal distribution and combined with the output of the CNN model. This joint vector acts as a conditioning input for the diffusion model, which samples from the distribution of high-fidelity cross-sections to refine the CNN output and produce a well-defined cross-section. b) The CNN encoder network used to approximate the large scale features of the high-fidelity output is adapted from the Residual in Residual Dense Network from the ESRGAN model \cite{wang2018esrgan}. c) The U-Net structure used in the diffusion model to predict the noise vector to be removed at each successive timestep of the denoising process. The model takes as input an intermediate image at timestep $t$ during the noising process, and outputs the noise to be removed at the specific timestep. 
}
    \label{fig:schematic}
\end{figure}

% \begin{figure}[h]
%     \centering
%     \includegraphics[width=\textwidth]{figures/Slide2.png}
%     \caption{    The unconditional diffusion process framework. During the training process, a sample high resolution image from the target dataset has noise iteratively added to it over a series of timesteps until the image is effectively isotropic gaussian noise. During this process, the network denoted $p_\theta$ is trained to be able to predict the amount of noise added to form the next timestep, based on the current timestep. In the inference step, a randomly sampled Gaussian noise vector can be provided to the model, where it will predict the corresponding amount of noise to remove at each timestep, Thus, a synthetic image can be sampled from the empirical distribution defined by the dataset by iteratively denoising a random vector.
% }
%     \label{fig:diffusion}
% \end{figure}
% \begin{figure}[htbp!]
%     \centering
%     \includegraphics[width=\textwidth]{figures/Slide2.png}
%     \caption{Diffusion Pipeline} 
%     \label{fig:pipeline}
    
% \end{figure}

\subsection{Diffusion Model}

% \begin{figure}[htbp!]
%     \centering
%     \includegraphics[width=\textwidth]{figures/Slide3.png}
  
%     \label{fig:diffusion}
% \end{figure}

\begin{figure}[htbp!]
    \centering
    \includegraphics[width=\textwidth]{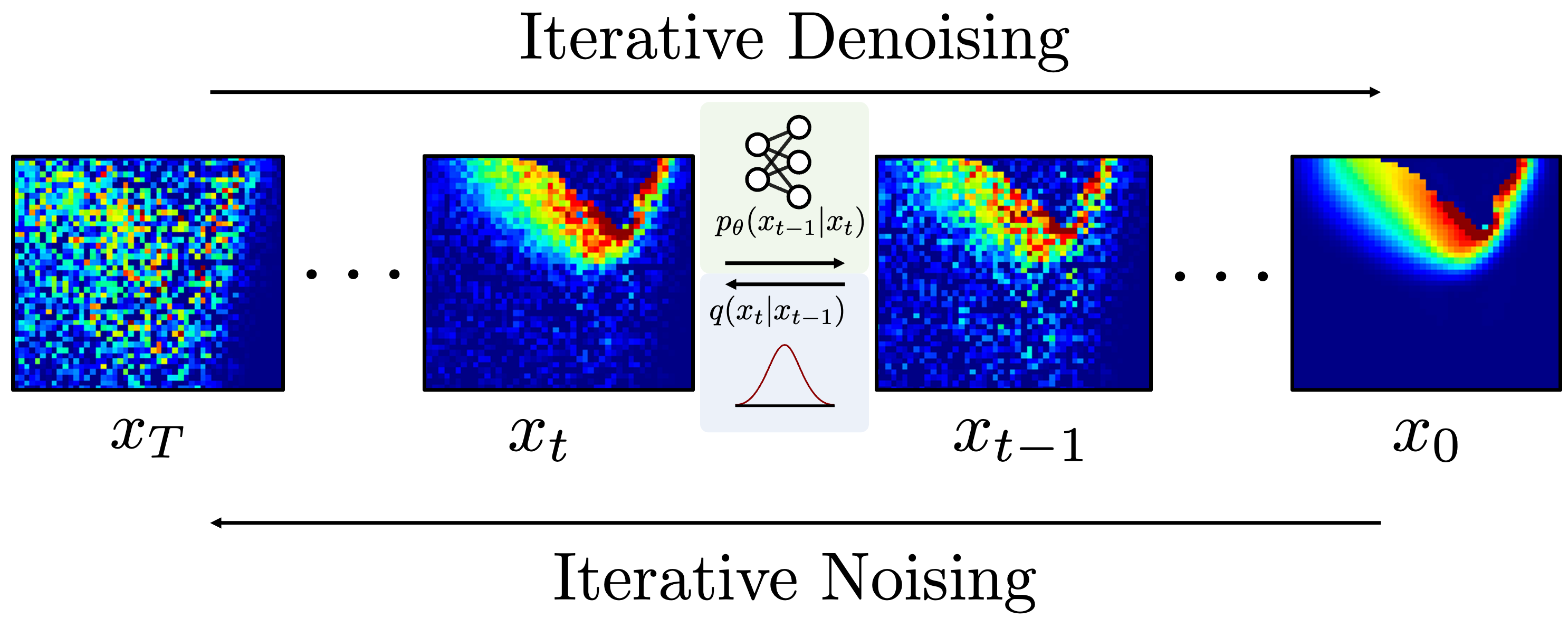}
    \caption{    The unconditional diffusion process framework. During the training process, a sample high fidelity cross-section from the target dataset has noise iteratively added to it over a series of timesteps until the image is effectively isotropic Gaussian noise. During this process, the network denoted $p_\theta$ is trained to be able to predict the amount of noise added to form the next timestep, based on the current timestep. In the inference step, a randomly sampled Gaussian noise vector can be provided to the model, where it will predict the corresponding amount of noise to remove at each timestep, Thus, a synthetic image can be sampled from the empirical distribution defined by the dataset by iteratively denoising a random vector.
}
    \label{fig:diffusion_schematic}
\end{figure}
Denoising Diffusion Probabilistic Models (DDPMs) are a class of generative models that create synthetic samples by iteratively transforming a Gaussian latent variable to the empirical distribution that defines the original dataset. DDPMs were originally designed for unconditional probabilistic sampling, to generate a synthetic datapoint by sampling arbitrarily from the parameterized data distribution, $p(x)$. The conditional setting extends the DDPM framework to enable sampling based on a specified input condition, modeling the data distribution $p(y|x)$, as opposed to the marginalized distribution $p(x)$.

DDPMs learn these potentially complex empirical data distributions by using a parameterized Markov chain. The Markov chain approach iteratively transforms a sampled Gaussian latent variable to a point in the empirical data distribution. To execute this, a forward diffusion process and a reverse denoising process are carried out during model training and inference respectively. The interaction of these two processes are represented visually in Figure \ref{fig:diffusion_schematic}, and additional detail about the algorithmic implementation is available in Ref. \cite{ho2020denoising}. In the forward process, a Markov chain adds Gaussian noise to a signal over $T$ timesteps. The amount of noise added at each timestep $t$ is parameterized by a predefined variance schedule, $\beta_{t}$.  Specifically, 

$$q(x_t|x_{t-1}) = \mathcal{N}(x_t; \sqrt{1 - \beta_{t}} x_{t-1}, \beta_{t} I)$$

The variance schedule defines how the scale of the noise applied to the image changes as a function of the diffusion timesteps. Typically, the noise scale will be smaller near the beginning of the forward process, and increase as the image becomes increasingly noisy. By reparameterizing the expression for $x_{t}$, it can be determined simply in terms of the variance schedule, the initial datapoint $x_{0}$, and the diffusion timestep $t$ with the following expression:

$$q(x_{t} | x_{0}) = \mathcal{N} (x_t ; \sqrt{\bar{\alpha_t}}  x_0 ,(1 - \bar{\alpha_t}) I) $$
where $\alpha_t = 1 - \beta_t$ and $\bar{\alpha_t} = \prod_{s = 1}^{t} \alpha_s $.

The reverse process begins with a data point assumed to be isotropic Gaussian noise, $x_{T}$, and iteratively de-noises the image by removing a predicted amount of noise, $\hat{\epsilon}_{t}$ at each time-step. This reverse process is given by 

$$p_{\theta} (x_{t-1} | x_{t}) = \mathcal{N} (x_{t-1}; \mu_{\theta} (x_t, t), \Sigma_{\theta} (x_t, t)) ) $$
where $p_\theta$ is a neural network. The network is trained to correctly predict the mean of the noisy data distribution conditioned on the noiseless image, $\mu (x_t| x_0)$.

% This has been shown to be equivalent to predicting $$ \mu_{\theta} = \frac{1}{\sqrt{\alpha_t}} \left ( x_t - \frac{\beta_t}{\sqrt{1-\bar{\alpha_t}}} \epsilon_\theta (x_t, t) \right ) $$ 

Therefore, the neural network representing $p_{\theta}$ is trained to optimize the difference between the predicted noise, $\epsilon_{\theta}$ and the added noise $\epsilon$ at each timestep. The structure of the $p_{\theta}$ network is shown in Figure \ref{fig:schematic}c.

During the inference process, a noise vector is sampled from the standard normal distribution, and undergoes the reverse process, transforming into a sample from the empirical data distribution, $p(x_0)$. In the conditional variant of this framework, the input signal that conditions the probability distribution is provided to the $p_{\theta}$ network alongside the noise vector. This restricts the training and inference process to learn the probability distribution $p(y | x)$, as opposed to the unconditional formulation, $p(y)$. Here, we follow the conditioning methodology outlined in \cite{li2022srdiff}, where a CNN encoder is used to transform the input low fidelity input to an embedding that captures the large-scale structure of the high fidelity target.

% Assuming the dataset consists of a set of input-output pairs, $\mathcal{D} = {x_i, y_i}_{i = 1}^{N}$,

% Diffusion probabilistic models use a Markov chain representation to model complex data distributions. Given an initial latent variable represented in a simple, analytical distribution, a parameterized Markov chain iteratively transforms the variable to a point in the original complex data distribtuion. This is done via two processes, a forward diffusion process, and a backward denoising process. In the forward process, a Markov chain gradually adds Gaussian noise to a signal until the signal is destroyed. The transition between a given state $t - 1$, and the state after a small amount of noise is added, $t$, is given by the distribution $q(\mathbf{x}_t| \mathbf{x}_{t-1}$). If this process occurs over a large number of timesteps, the transitions of the noise addition process can be parameterized by a neural network. This parameterization allows us to model the amount of noise to be removed at each timestep to obtain a clear image, $p_{\theta} (\mathbf{x}_{t-1} | \mathbf{x}_t)$.   
\subsection{Encoder Model}
\label{subsection:encoder}
To transform the low fidelity data into the mean of the high fidelity distribution conditioned on the low fidelity data, we make use of the Residual in Residual Dense architecture (RRDN), show in Figure \ref{fig:schematic}b . The RRDN, first proposed by Wang et al. (2018) \cite{wang2018esrgan} originally acted as the generator network in a Generative Adversarial Network (GAN) architectural approach to super-resolution, and consists of several hierarchical layers of residual blocks. Within each residual block, dense connections, residual learning and contiguous memory mechanisms enhance information and gradient flow. The structure of the model is provided in Figure \ref{fig:schematic}, where each basic block is a residual-in-residual block. The operations take place mostly in the reduced order low-resolution space to reduce the computational demands of the training process. Following the completion of the residual-in-residual layers of the neural network, multiple up-sampling operations are applied to shift the data into the high-resolution space. This model is trained using an $L_1$ loss function. Once the model has been trained to minimize the $L_1$ loss between the input-output pairs to an adequate level, the weights are frozen during the inference process and the diffusion training process. During the inference process, the output of the model prior to the last convolutional layer, a [64 $\times$ W $\times$ H] size tensor, where W and H denote the size of the high-resolution image, is extracted as the encoder output data. Therefore, the encoder model is able to provide a more principled conditioning than bicubic interpolation.

\subsection{Model Architecture}

The overall model pipeline consists of an encoder model that predicts the approximate mean of the empirical distribution of potential high fidelity results given  a low fidelity input, followed by a diffusion process to fine tune and sample this distribution. Specifically, we first feed the low fidelity input image to the CNN encoder model, which is an RRDN generator architecture as described in Section \ref{subsection:encoder}. This model is trained based on an objective that minimizes the $L_1$ loss between the produced image and the target image. Once this model has been trained, the output of the model before the last convolution layer is combined with the noise sampled to input to the diffusion model, to form an input that provides conditioning information for the diffusion  model denoising process. At the conclusion of the denoising process, the recovered image is the final model output. The overall inference process for the conditional diffusion model is shown in Figure \ref{fig:schematic}a and Figure \ref{fig:pipeline}b. 

\subsection{Dataset Generation}
The simulation data for this work is generated by simulating bare plate, single track runs of Stainless Steel Grade 316L (SS316L) and Ti-6Al-4V at varying process parameters. Each simulation is conducted with multiple mesh sizes to extract low-fidelity and high-fidelity information for the model training. Specifically, 295 simulations are conducted using SS316L at a $10 \mu m$ mesh element size for high-fidelity training data and a $20 \mu m $ mesh element size for low-fidelity training data. Forty simulations are conducted using Ti-6Al-4V, with a $5 \mu m $ mesh element used for the high-fidelity training data, and 20 $\mu m $ mesh elements used for low-fidelity training data. The process parameters for the SS316L case are chosen to span the conduction mode and keyhole mode regimes of melt pool behavior, while the process parameters for the Ti-6Al-4V case are selected to prioritize simulations that contain complex, non-deterministic keyhole behavior. Each simulation describes the behavior resulting from a 100$\mu m$ diameter moving laser impacting a solid build plate spanning $ 1 mm $ in the laser plane of travel, $0.6 mm$ perpendicular to the plane of travel, with a thickness of $0.4 mm$. The laser travels along the build plate for 500 $\mu s$ with a specified constant velocity, $V$ and power, $P$, varying from 150 mm/s to 1400 mm/s and 100 W to 500 W respectively. The 3D simulation output fields are captured at intervals of $5 \mu s$ during the laser melting process. From the transient 3D temperature field produced by the simulation, a 2D cross-section is taken along the laser plane of travel. This cross-section is cropped to focus on the melt pool, and travels with the laser to ensure the melt pool is consistently centered within the image. Specifically, this cross section is a $320\mu m $ $\times$ 320$\mu m$ area horizontally centered on and extending downwards from the impact point of the laser with the substrate. The material parameters of the SS316L simulations were validated through comparison with the geometrical properties of experimental single-bead melt tracks \cite{myers2023high}, while the material parameters of the Ti-6Al-4V simulations were validated through comparison with experimentally reported literature values \cite{cunningham2019keyhole}. Direct comparisons of the quantities studied are available in \ref{sec:sample:appendix} and in Ref. \cite{hemmasian2023surrogate}.

The computational fluid dynamics (CFD) software FLOW-3D is used to simulate the melting behavior \cite{FLOW-3D}. To do so, FLOW-3D solves the following coupled PDEs describing the fluid flow and heat transfer phenomena of the melt pool.

\begin{equation}
    \nabla \cdot (\rho \vec{v}) = 0
\end{equation}

\begin{equation}
    \frac{\partial \vec{v} }{\partial t} + (\vec{v} \cdot \nabla) \vec{v} = - \frac{1}{\rho}\nabla \vec{P} + \mu \nabla^2 \vec{v} + \vec{g}(1-\alpha (T-T_m))
\end{equation}

\begin{equation}
\frac{\partial h}{\partial t} + \left ( \vec{v} \cdot \nabla \right ) h = \frac{1}{\rho}\left( \nabla \cdot k \nabla T \right) 
\end{equation}

where $\vec{g}$ is gravity, $\alpha$ is the coefficient of thermal expansion, $\vec{P}$ is pressure,  $\rho$ is the density, $\vec{v}$ is velocity, $h$ is the specific enthalpy and $k$ represents heat conductivity. Additionally, the Boussinesq approximation models the variation of density with temperature. These equations are simulated on a Cartesian mesh, with a Volume-of-Fluid (VOF) framework to represent the interface between the plate and the plate surroundings \cite{hirt1981volume}. Additional considerations are applied in order to account for the phase change phenomena induced during the laser melting process. The thermal properties of each material simulated are temperature dependent, and their relationships with temperature are referenced from \cite{mills2002recommended}.  In order to reduce the computational expense of the simulation, the fluid dynamics of the gas phase above the substrate are not simulated explicitly, but are approximated with relationships defining the pressure exerted on the free surface, in addition to the  heat and mass transfer caused by material evaporation. The heat input provided by a laser of radius $r_0$ irradiating the surface with power $P$ is parameterized by a Gaussian distribution, described in Equation \ref{eq:laser_input}. 
\begin{equation}
q = \frac{P}{r_0^2 \pi}\exp \left \{ - \left ( \frac{\sqrt{2} r}{r_0} \right )^2 \right \}
\label{eq:laser_input}
% \end{equation
\end{equation}

In high energy density cases, the substrate vaporizes, causing a recoil pressure that creates a keyhole-shaped vapor cavity within the melt pool. This cavity increases the effective absorptivity of the laser from the nominal flat surface value due to reflection effects \cite{trapp2017situ, ye2019energy}. Therefore, a Fresnel multiple reflection model is used to describe the effects of the laser incidence angle on the effective absorptivity. This absorptivity is based on a material dependent property $\epsilon$, set to 0.25 for Ti-6Al-4V, and 0.15 for SS316L in this work. The absorptivity of a laser ray impacting with incidence angle $\theta$ is given by
\begin{equation}
A = 1 - \frac{1}{2}\left( \frac{1+(1-\epsilon \cos \theta )^2}{1+(1+\epsilon \cos \theta )^2}  + \frac{\epsilon^2 - 2 \epsilon \cos \theta + 2 \cos^2 \theta}{\epsilon^2 + 2 \epsilon \cos \theta + 2 \cos^2 \theta}\right  )
\end{equation}

The recoil pressure exerted on the free surface by the vaporization process is described in Equation \ref{eq:pressure}, where $\Delta H_v$ is the heat of vaporization, and $a$ is the accommodation coefficient representing the ratio of mass exchange between phases as a fraction of the maximum mass exchange thermodynamically possible. Also, $\gamma$ is the ratio of the constant pressure and constant temperature specific heats, $C_v^{vap}$ is the constant volume heat capacity of the metal vapor, and $P_V$,  $T_V$ are the saturation pressure and temperature respectively. 
\begin{equation}
    \label{eq:pressure}
    P_s = a P_{V} \exp \left \{  \frac{\Delta H_v}{(\gamma - 1)c_v T_v} \left ( 1 - \frac{T_v}{T}\right ) \right \}
\end{equation}

 The evaporative mass loss, $\dot{M}$ is approximated according to Equation \ref{eq:evapmassloss}, where $P_p$ is the specified pressure of the void phase, $\bar{T}$ is the average temperature of the liquid along the free surface, and $R$ is the gas constant.

\begin{equation}
    \label{eq:evapmassloss}
    \dot{M} = a \sqrt{\frac{1}{2 \pi R \bar{T}}}(P_l^{sat} - P_p)
\end{equation}

\section{Results and Discussion}

\label{sec:sample3}
To evaluate the effectiveness of the model, we design a 4$\times$ upscaling task (20$\mu m$ to 5$\mu m$) and a 2$\times$ upscaling task (20$\mu m$ to 10$\mu m$) based on the collected Ti-6Al-4V and SS316L datasets respectively.
\subsection{4$\times$ upscaling task}

\begin{figure}[htbp!]
    \centering
    \includegraphics[width=\textwidth]{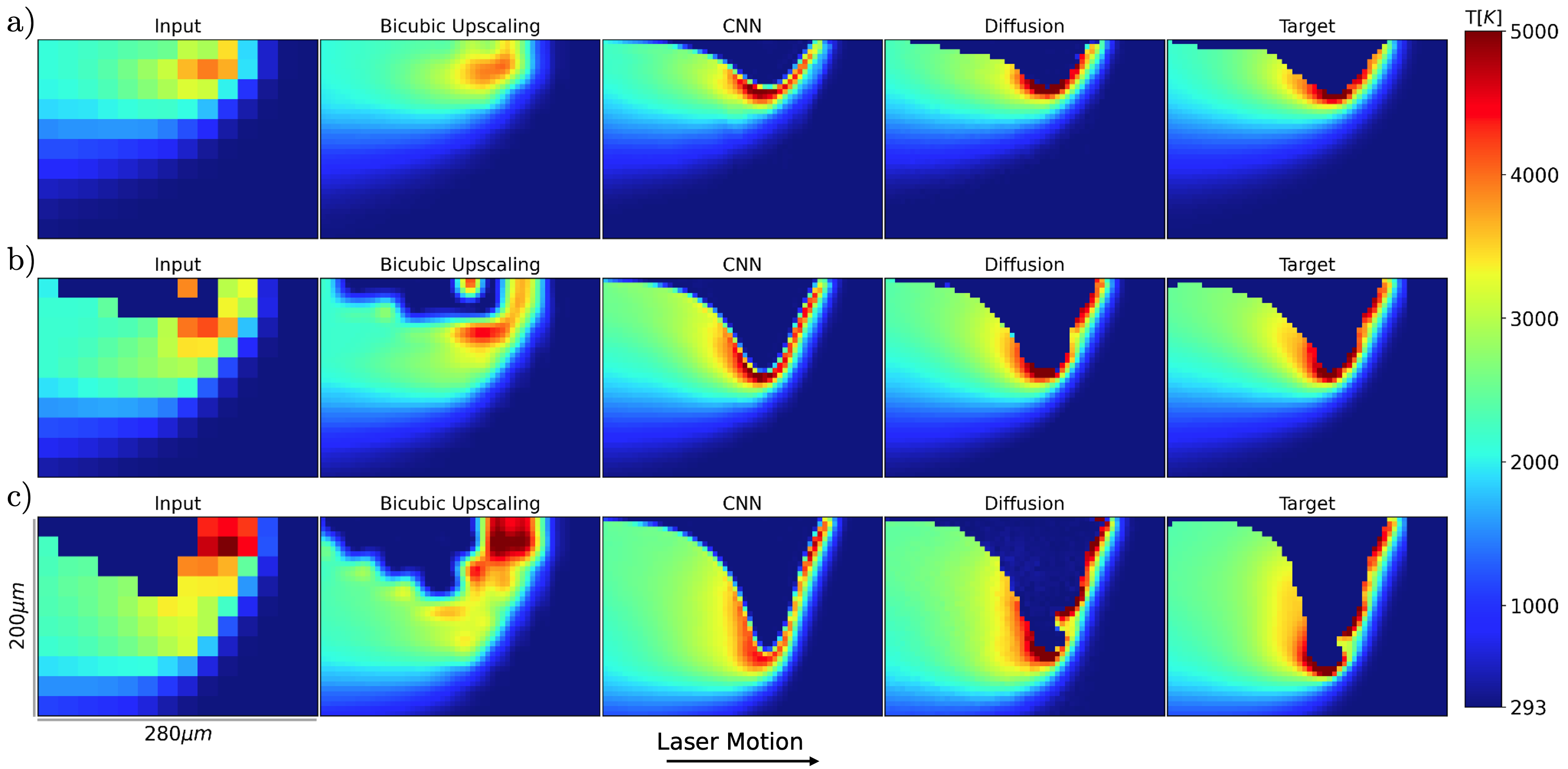}
    \caption{Diffusion based upscaling of the temperature field  for three different processing parameter combinations.  The input low fidelity data, in addition to the bicubic upscaling, CNN, and diffusion predictions are shown for a) Power (P) = 156 W, Velocity (V) = 700 mm/s. b) P = 260 W, V = 900 mm/s. c) P = 364 W, V = 900 mm/s. The cross-sections shown here are taken along the laser plane of travel, at the mesh element closest to center of the melt track.}
    \label{fig:diffusion_results}
\end{figure}
% \FloatBarrier

\begin{figure}[htbp!]
    \centering
    \includegraphics[width=\textwidth]{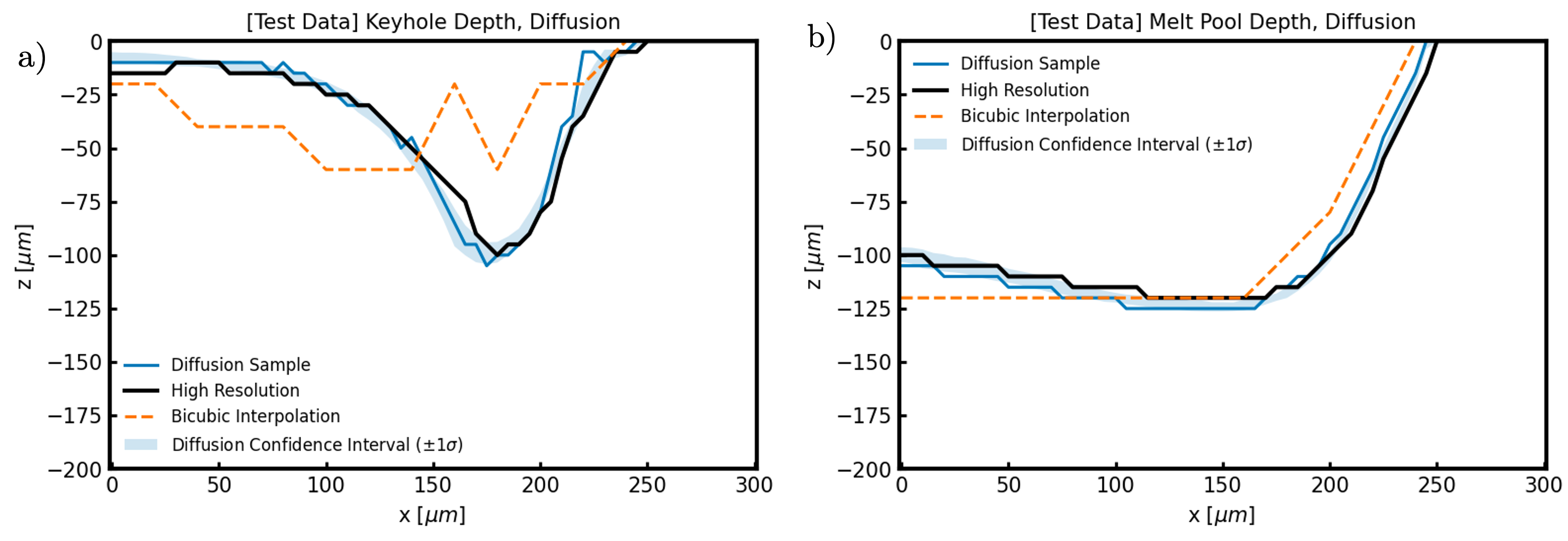}
   \caption{a) The vapor cavity produced by bicubic interpolation, diffusion upscaling compared to the high fidelity target image, at P = 260 W, V = 900 mm/s, and t = 495 $\mu s$ into the trajectory. The cross-sections shown here are taken along the laser plane of travel, at the mesh element closest to center of the melt track. The melt pool surface profiles produced by bicubic interpolation, diffusion upscaling compared to the high fidelity target image, at P = 260 W, V = 900 mm/s, and t = 495 $\mu s$ into the trajectory. The cross-sections shown here are taken along the laser plane of travel, at the mesh element closest to center of the melt track.}
    \label{fig:ti64diffusion_contour}
\end{figure}

\subsubsection{Training Details}

In this experiment, the dataset consists of 40 simulations with mesh elements of size 5 $\mu m$ and 40 simulations with 20 micron $\mu m$ mesh elements. Each simulation describes a single melt track on bare plate Ti-6Al-4V, at conditions ranging from P = 120 W to P = 520 W, and V = 400 mm/s to 1200 mm/s.  The encoder CNN model is trained for 300 epochs with a learning rate of 1E-4, using the Adam optimizer. Data augmentation is applied by flipping 25\% of the samples observed during training about the x-axis, emulating a laser traveling in the opposite direction in the same plane. These augmentations aim to reduce overfitting and improve the model performance on unseen data by increasing the effective number of unique samples that the model is trained on.  Following the completion of the CNN encoder training process, the CNN weights are frozen, and the diffusion model is trained. This training process takes place using the pre-trained encoder to condition the diffusion prediction on the low-fidelity input. The diffusion model is also trained for 300 epochs to minimize the Huber loss of the predicted noise distribution against the sampled noise distribution at each time step of the process. Here, the diffusion model uses 1000 noising timesteps with a linearly interpolated variance schedule between 0.0001 and 0.02 in order to train the model and produce new samples.

A modified sampler, taken from \cite{song2020denoising}, accelerates the inference process by replacing the default Markovian sampler originally proposed in \cite{ho2020denoising} with a non-Markovian deterministic sampler. The original formulation of the sampler requires consecutive iteration through \textit{T} timesteps to denoise an image from isotropic Gaussian noise to a coherent sample. The modified sampler no longer requires iteration through all \textit{T} timesteps, rather, a subset of these timesteps can be used for the sampling process. Here, we iterate the modified sampler once every 50 timesteps of the diffusion denoising process, for a 50$\times$ reduction in inference time compared to the original sampler.  Before the input is passed to the model, a cell-wise re-normalization is applied in order to ensure that the magnitude of the values  used during the training process is centered in the range [-1,1]. Specifically, the mean and standard deviation value for each cell in the cross-section in the training dataset. Once these values are computed, the data is normalized by subtracting the mean and dividing by the standard deviation.  By re-normalizing the data, the data now fulfills the standard normal distribution assumption of the diffusion training process. To benchmark the model ability to generalize to unseen inputs, we divide the generated simulations into a training set, a validation set, and a test set. 75\% of the simulations are used as the training set for the model, while 10\% of the simulations are reserved to tune model performance. Finally, 15\% of the simulations are reserved to evaluate the tuned model. The distribution of simulations used in each partition is represented in Figure \ref{fig:datadescription}.
\subsubsection{Model Benchmarks}

\begin{figure}[htbp!]
    \centering
    \includegraphics[width=\textwidth]{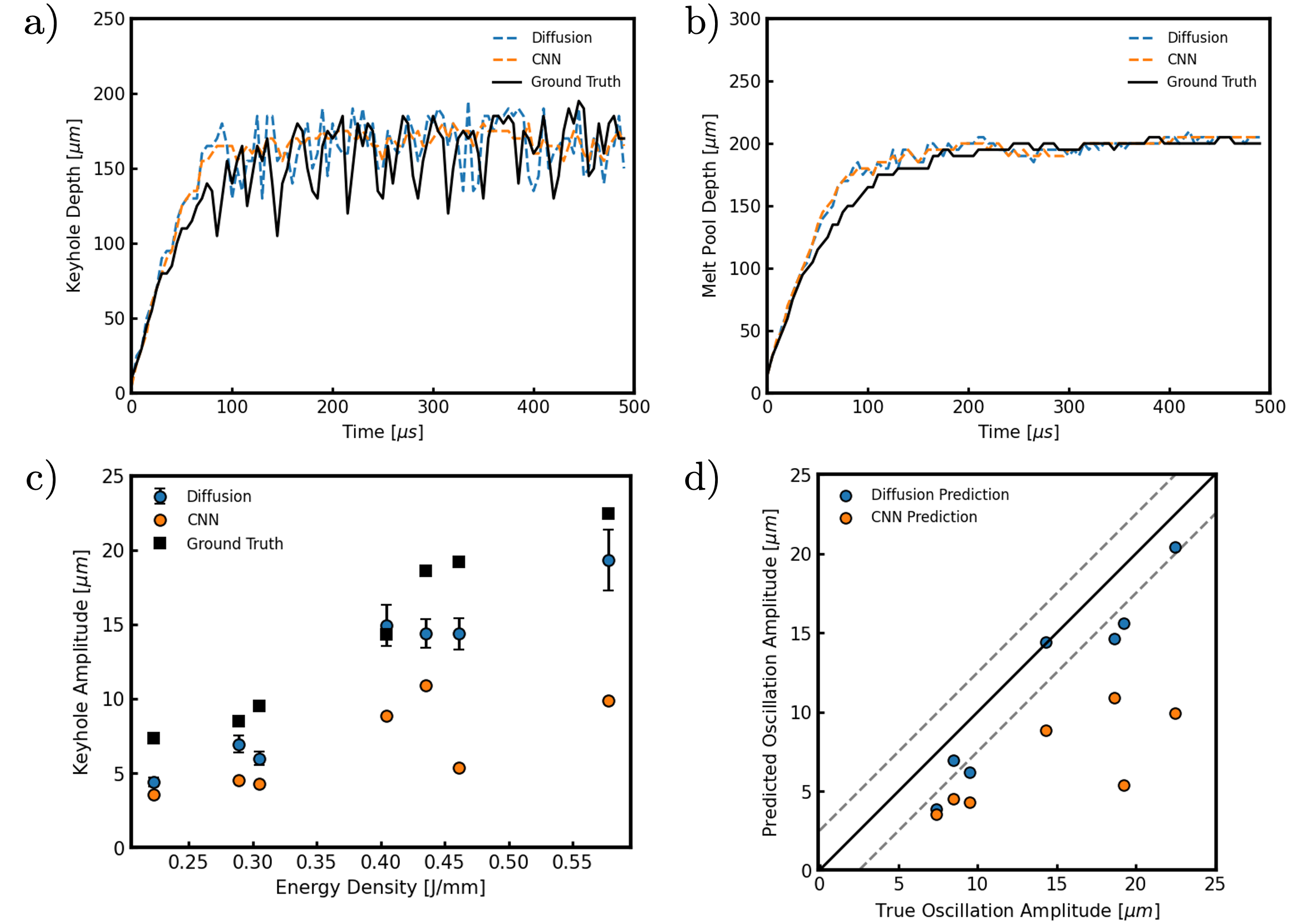}
  
    \caption{
   a) The melt pool depth for the diffusion and CNN-based predictions compared to the ground truth simulated melt pool depth over time, for an unseen simulation (P = 415 W, V = 900 mm/s) in the Ti-6Al-4V test set. b) The depth of the keyhole cavity for the diffusion and CNN-based predictions compared to the ground truth simulated keyhole depth over time for the same simulation. The diffusion model produces samples that have the same degree of variability as the ground truth data, despite the inability for the low-fidelity input data to effectively represent the keyhole structure.  c) The keyhole oscillation amplitude as a function of the input laser energy density for the ground truth data, the diffusion model output, and the CNN model. The errorbars for the diffusion model represent the variation present in ten different generations of the diffusion model under the same low-resolution input. d) The amplitude of the predicted keyhole oscillation for the CNN and diffusion model. The dotted lines indicate a 2.5 $\mu m$ over-prediction or under-prediction  from the optimum value, comparable to the uncertainty caused by the 5 $\mu m$ mesh size.   }
    \label{fig:diffusionfreqamp}
   
\end{figure}

First, we evaluate the image quality of the reproduced samples. For each simulation, we extract the 2D cross-section along the direction of laser travel at each timestep. Next, we benchmark the performance of several upscaling methods for recreating a feasible high fidelity temperature field from the low fidelity input. Specifically, we evaluate the performance of bicubic upscaling, the performance of a deterministic CNN model, and finally, the performance of the proposed diffusion model. In Figure \ref{fig:diffusion_results}, the bicubic upscaling attempts to upscale the low fidelity data, via interpolation. However, due to the large-scale discrepancies in the low fidelity data distribution caused by the unresolved physics in the model, the interpolation of these values does not yield a useful output. Qualitatively examining the CNN model we observe that the output is able to resolve the large scale features of the melt pool. For instance, the melt pool temperature distribution is closely matched between the CNN output and the ground truth high fidelity output, as well as the general shape of the keyhole vapor depression. However, the small-scale details present in the high fidelity ground truth are not visible in the output of the CNN model. The diffusion model resolves these small scale features by modeling the empirical distribution of possible high fidelity outputs given the low fidelity input data. With this sampling process, the diffusion model is able to create a feasible reconstruction of the small scale features of the high fidelity input. Resolving these small-scale features, such as the protrusions and localized areas of high temperature on the keyhole wall, can also provide an indication of other defect formation mechanisms outside of keyhole collapse. For example, Zhao et al. demonstrated spatter formation through bulk explosion of high-temperature ligaments on the front keyhole wall, formed based on the recoil pressure interactions with the melt pool free surface \cite{zhao2019bulk}.

Following this qualitative benchmark of the melt pool image reconstruction, we evaluate the reconstruction of the melt pool with quantitative metrics. Due to the influence of the melt pool and keyhole geometry on the mechanical properties of the final printed part, we select the melt pool depth and keyhole depth as evaluation metrics for the overall model. The contour lines for a set of predicted target and reconstructed melt pools are shown in Figure \ref{fig:ti64diffusion_contour}, showing direct improvement over bicubic interpolation.
% These issues increase the potential space of possible high resolution vapor structures given a low resolution input.  

Next, we extract more sophisticated metrics of the melt pool behavior to further evaluate the performance of the diffusion model. Ren et al. (2023) observe two fundamental modes of the fluctuation of the keyhole, intrinsic and peturbative oscillations, that can serve as an indicator of the tendency for a keyhole vapor depression to collapse and trap pores underneath the surface \cite{ren2023machine}. These unstable keyholes are characterized by stochastic protrusions on the surface of the keyhole wall. To benchmark the ability of the model to recreate the variability in the keyhole that can lead to pore formation, we extract the amplitude of the observed temporal  keyhole fluctuation. To do so, we compute the standard deviation of the keyhole depth with respect to time in the quasi-steady state region to obtain the average amplitude. By plotting these quantities as a function of the energy density of the laser heat input, we can see that the tendency for the amplitude of variation to increase with the heat input is captured in the high fidelity simulation, the output of the CNN model, and the output of the diffusion model. However, the CNN model is unable to capture the magnitude of the keyhole structure oscillation as effectively as the diffusion model output. This behavior is demonstrated in Figure \ref{fig:diffusionfreqamp}c and Figure \ref{fig:diffusionfreqamp}d.

Finally, to examine the image reconstruction quality of the diffusion pipeline, the $L_1$ loss between the upscaled result and the high fidelity result is observed. 
%$$ \mathrm{PSNR} = 10 \log_{10} \left(\frac{(MAX_I)}{\mathrm{MSE}}\right)$$

% $$\mathrm{MSE} = \frac{1}{mn} \sum_{i=0}^{m-1} \sum_{j=0}^{n-1} [I(i,j)-K(i,j)]^2$$
% %
 The $L_1$ loss is calculated as $$L_{1}(\hat{y}, y) = \frac{1}{n} \sum_{i=1}^{n} \left|\hat{y}_{i} - y_{i}\right|$$ and measures the discrepancy between the temperature of the reconstructed and target images. Accordingly, there is a clear increase in the quality of the image in the high fidelity image, when compared to the bicubic upscaling of the low fidelity sample. To quantify the performance of the upscaling model for applications directly relevant to Laser Powder Bed Fusion, we define two metrics, the MP-MAE and the VC-MAE to quantify the error in the produced melt depth and vapor cavity respectively. Specifically, we find the deepest points along the melt pool and vapor cavity in both the predicted and ground truth for each image in the dataset, and calculate the mean average error of the predicted depths compared to the ground truth depths. These results are reported in Table \ref{table:metrics}. While the MAE of the CNN model is lower than the MAE of the diffusion model, this is due to the CNN only representing the mean of the possible high-fidelity output, as opposed to sampling from the distribution of the high-fidelity output. Although this yields a lower MAE in the CNN model, the CNN temperature field fails to establish a distinct boundary for the free surface.

\begin{figure}[htbp!]
    \centering
    \includegraphics[width=\textwidth]{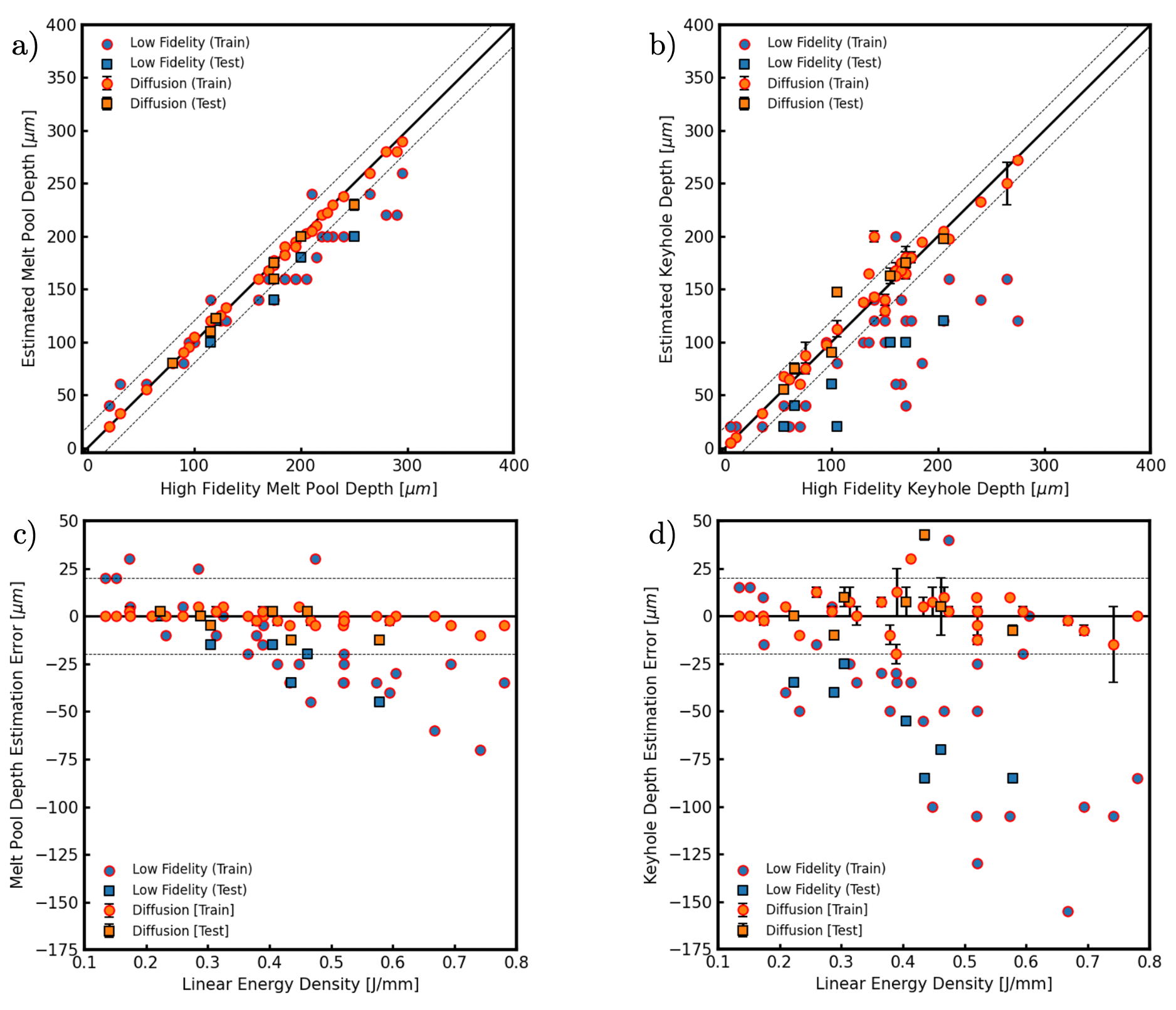}
  
    \caption{
   a) The melt pool depth for the low-fidelity and diffusion estimations compared to the high fidelity melt pool depth for simulations in the Ti-6Al-4V dataset at $t = 395 \mu s$. b) 
The keyhole depth for the low-fidelity and diffusion estimations compared to the high fidelity keyhole depth for simulations in the Ti-6Al-4V dataset at $t = 395 \mu s$. c) The melt pool depth estimation error for the low-fidelity and diffusion models compared to the linear energy density for simulations in the Ti-6Al-4V dataset at $t = 395 \mu s$. d) The keyhole depth estimation error for the low-fidelity and diffusion models compared to the linear energy density for simulations in the Ti-6Al-4V dataset at $t = 395 \mu s$.
The errorbars for the diffusion model represent the standard deviation across five independent samples from the diffusion model using the same low-resolution input. The dotted lines indicate a 20 $\mu m$ over-prediction or under-prediction  from the high-fidelity model value.}
    \label{fig:lrhrcomparison}
   
\end{figure}

The coarse resolution of the low-fidelity model sacrifices the ability to resolve smaller-scale features of the melt pool for faster run-time. Here, we quantitatively examine the discrepancy between the low-fidelity and high-fidelity model predictions based on the melt pool dimensions, and evaluate the ability of the diffusion model to recapture the unresolved behavior. To do so, we examine the melt pool depth and keyhole depth at a specific simulation time step across a range of laser energy densities for the low-fidelity model, the high-fidelity model and the diffusion model output (Figure \ref{fig:lrhrcomparison}a, \ref{fig:lrhrcomparison}b). From this analysis, we observe that the complexity of the melt pool behavior and correspondingly, the discrepancy between the low-fidelity and high-fidelity simulation behavior grows with the laser energy density.

 This effect is again likely due to the rapidly fluctuating keyhole cavity that forms at higher energy densities, which is difficult to resolve on a coarse mesh. However, this divergence is no longer observed in the predictions of the diffusion model (Figure {\ref{fig:lrhrcomparison}}). In order to ensure that this behavior generalizes to new simulations under arbitrary process conditions, we compare the accuracy of the melt pool and keyhole depth predictions for simulations not contained in the training set.

In both simulation cases included in the training set and unseen cases drawn from the withheld test set, the diffusion model produces predictions that closely align with the high-fidelity melt pool and keyhole depths, generally falling within a ± 20 $\mu m$ margin as shown in Figure {\ref{fig:lrhrcomparison}}c and Figure {\ref{fig:lrhrcomparison}}d. The similarity between the test accuracy and training accuracy demonstrates that this model can be used to interpolate within the process parameter space, enabling efficient  parameter sweeps with reduced computational time.

\begin{table}[htbp!]
\caption{Metrics evaluating the performance of the  Diffusion model on the 4$\times$ upscaling task.}
\centering
\begin{tabular}{@{}llll@{}}

\toprule
         Upsampling Method            & MAE [K]  & MP-MAE [$\mu m$] & VC-MAE [$\mu m$] \\ \midrule
Bicubic Upsampling   &  241      & 22.0       & 50.6        \\
CNN Upsampling       &  65.1        &    3.47  &   10.9     \\
Diffusion Upsampling &  67.3     &  3.05      &  12.1      \\ \bottomrule
\label{table:metrics}
\end{tabular}
\end{table}

%[Data availability benchmark]

%[Augmentation discussion]

%[Cooling Rate Benchmark]

% \subsection{Off-Nominal Process Parameter Prediction}
% To-do

% \subsection{Interpolation between timesteps}
% To-do

\subsection{2$\times$ upscaling task}
\subsubsection{Training Details}
A similar process is repeated for the $2\times$ upscaling task perfomed on the SS316L dataset. To facilitate this experiment, we create a dataset of 295  10-micron and 20-micron resolution single-bead simulations of SS316L at process parameters ranging from V = 100 mm/s to V = 950 mm/s, and P = 50 W to 490 W. Again, the encoder CNN model is trained for 300 epochs with a learning rate of 1 $\times 10^{-3}$ and a horizontal flip data augmentation is randomly applied with a probability of 25\%. In order to improve the convergence of the model training process, we also renormalize the values to lie within the range [-1, 1] by computing the cell-wise mean and standard deviation of the temperature fields in the dataset.

To evaluate the performance of the model on this task, we again examine the melt pool dimensions and the MAE of the prediction task on the overall temperature field. The results of these evaluation metrics are reported in Table \ref{table:ss316l_metrics}.

\subsubsection{Model Benchmarks}

\begin{figure}[htbp!]
    \centering
    \includegraphics[width=\textwidth]{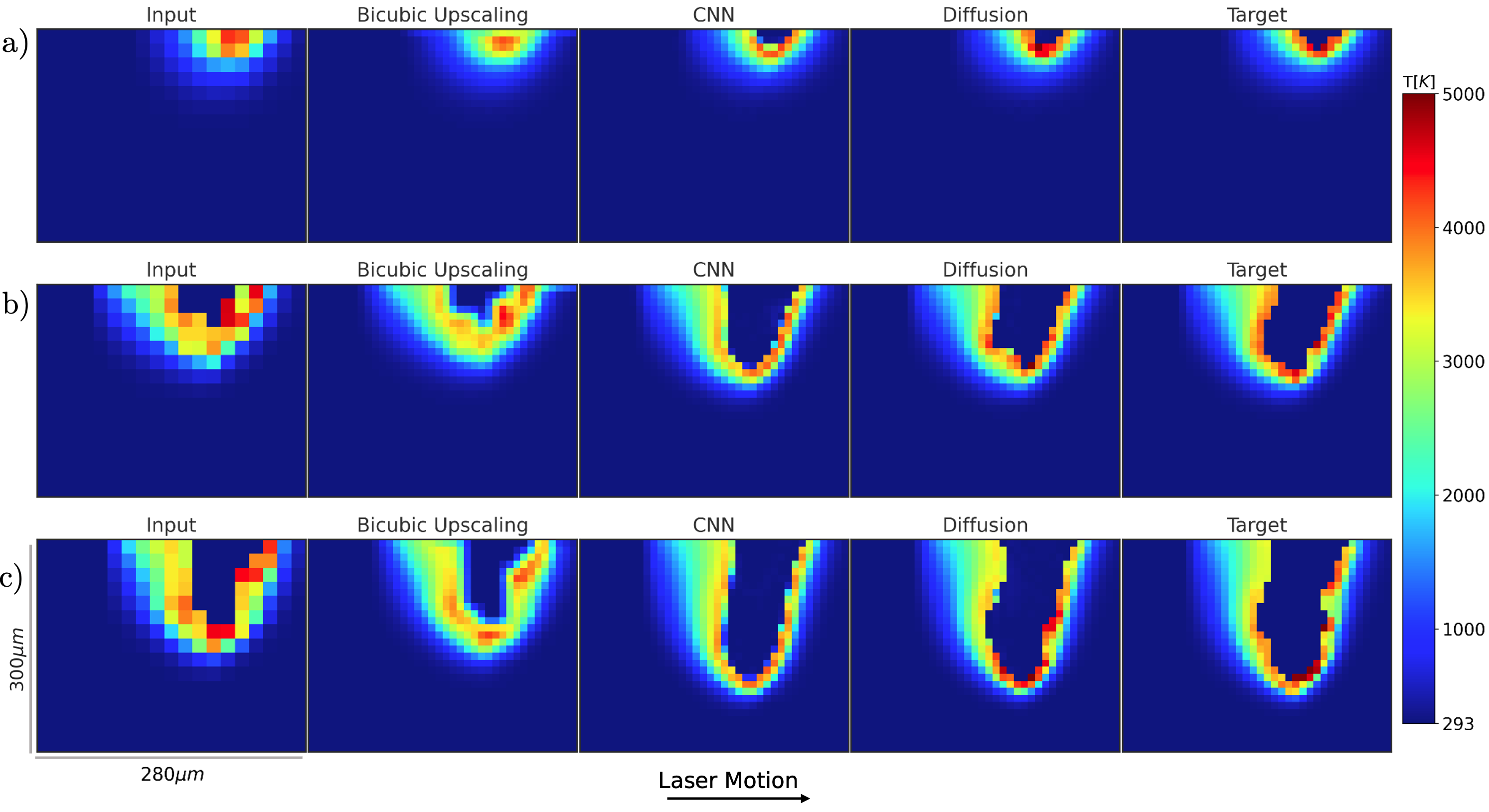}
   \caption{Diffusion based upscaling of the temperature field  for three different processing parameter combinations in the SS316L dataset at $t = 185 \; \mu s$. The input low fidelity data, in addition to the bicubic upscaling, CNN, and diffusion predictions are shown for a) P = 130 W, V = 250 mm/s. b) P = 310 W, V = 550 mm/s. c) P = 430 W, V = 500 mm/s.}
    \label{fig:diffusion_cross_section_ss316l}
\end{figure}

We evaluate the image quality of the reconstructed images for the 2$\times$ upscaling task by comparing the performance of bicubic interpolation, the upsampling CNN model, and the implemented diffusion model. A qualitative comparison between the three methods is made in Figure \ref{fig:diffusion_cross_section_ss316l}, demonstrating the superior ability of the diffusion model to create a reasonable free surface boundary compared to the CNN model.

\begin{figure}[htbp!]
    \centering
    \includegraphics[width=\textwidth]{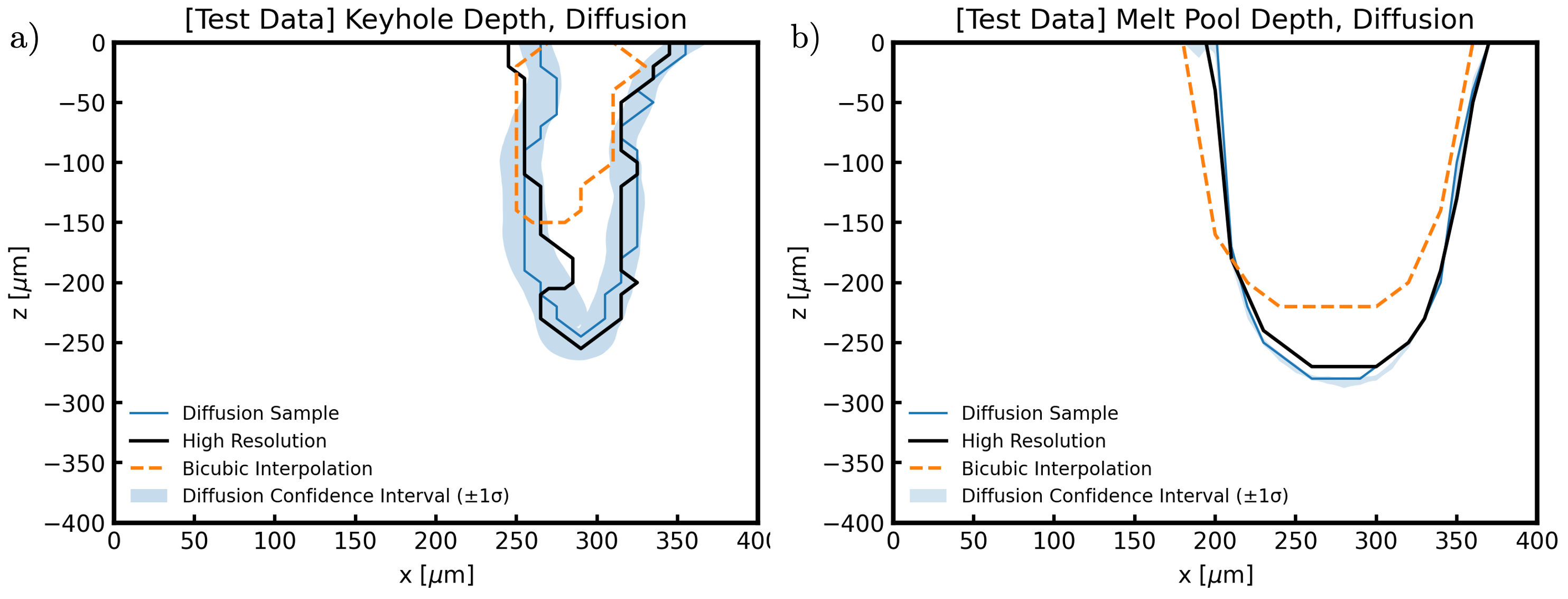}
   \caption{The vapor cavity (a) and melt pool surface profiles (b) predicted by bicubic interpolation of the low-fidelity model and diffusion upscaling of the low-fidelity model, compared to the high fidelity target sample. This comparison is shown for a sample from the SS316L dataset, P = 280 W, V = 100 mm/s at $t$=405 $\mu s$ into the scanning trajectory. }
    \label{fig:diffusion_contour}
\end{figure}

Similar to the performance observed in the $4 \times$ upsampling scenario, the bicubic interpolation fails due to the large-scale discrepancies between the low fidelity data and high-fidelity data. The CNN model is once again able to approximate the general shape of the melt pool, but is unable to create plausible high-level details due to the tendency of the CNN to average over the space of possible output images. The diffusion model is able to sample from this distribution, creating qualitatively reasonable high-level melt pool features. This is most notable in the ability of the diffusion model to recreate the sharp boundaries differentiating the melt pool from the keyhole void space. The distribution of feasible melt pool profiles and keyhole profiles based on a given low-fidelity input is shown in Figure \ref{fig:diffusion_contour} for a sample cross-section in the dataset.

\begin{figure}[htbp!]
    \centering
    \includegraphics[width=\textwidth]{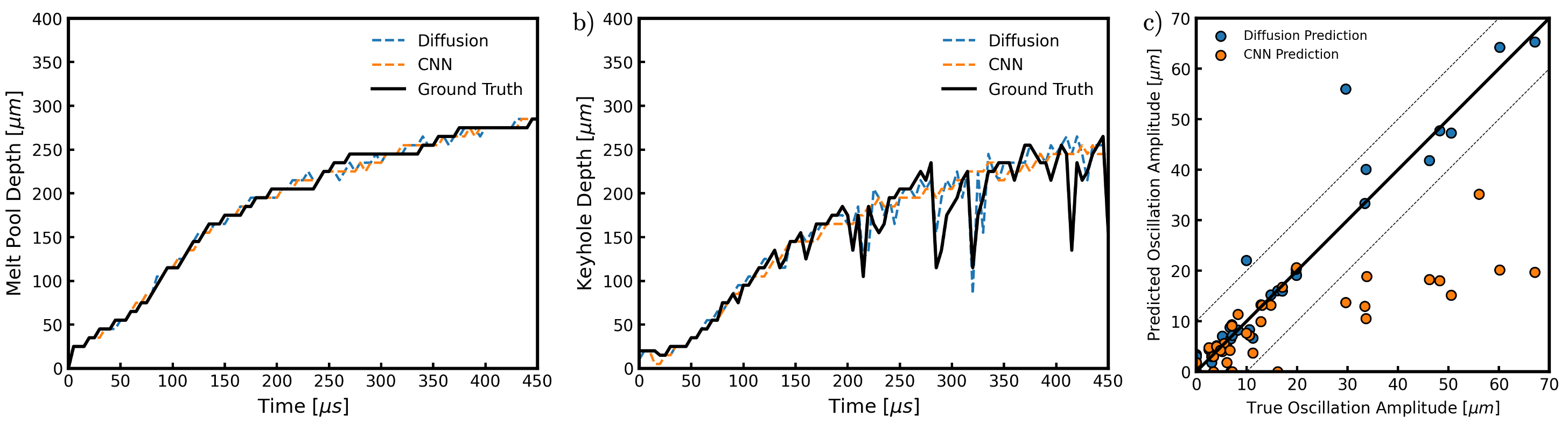}

    \caption{a) The melt pool depth for the diffusion and CNN-based predictions compared to the ground truth simulated melt pool depth over time, for a simulation (P = 280 W, V = 250 mm/s) in the SS316L test set. b) The depth of the keyhole cavity for the diffusion and CNN-based predictions compared to the ground truth simulated keyhole depth over time for an unseen simulation in the test set. The diffusion model produces samples that have the same degree of variability as the ground truth data, despite the inability for the low-fidelity input data to  resolve the keyhole structure.  c) The amplitude of the predicted keyhole oscillation for the CNN and diffusion model. The dotted lines indicate a 10 $\mu m$ over-prediction or under-prediction  from the optimum value.  }
    \label{fig:ss316ldiffusionfreqamp}
\end{figure}

\begin{figure}[htbp!]
    \centering
    \includegraphics[width=\textwidth]{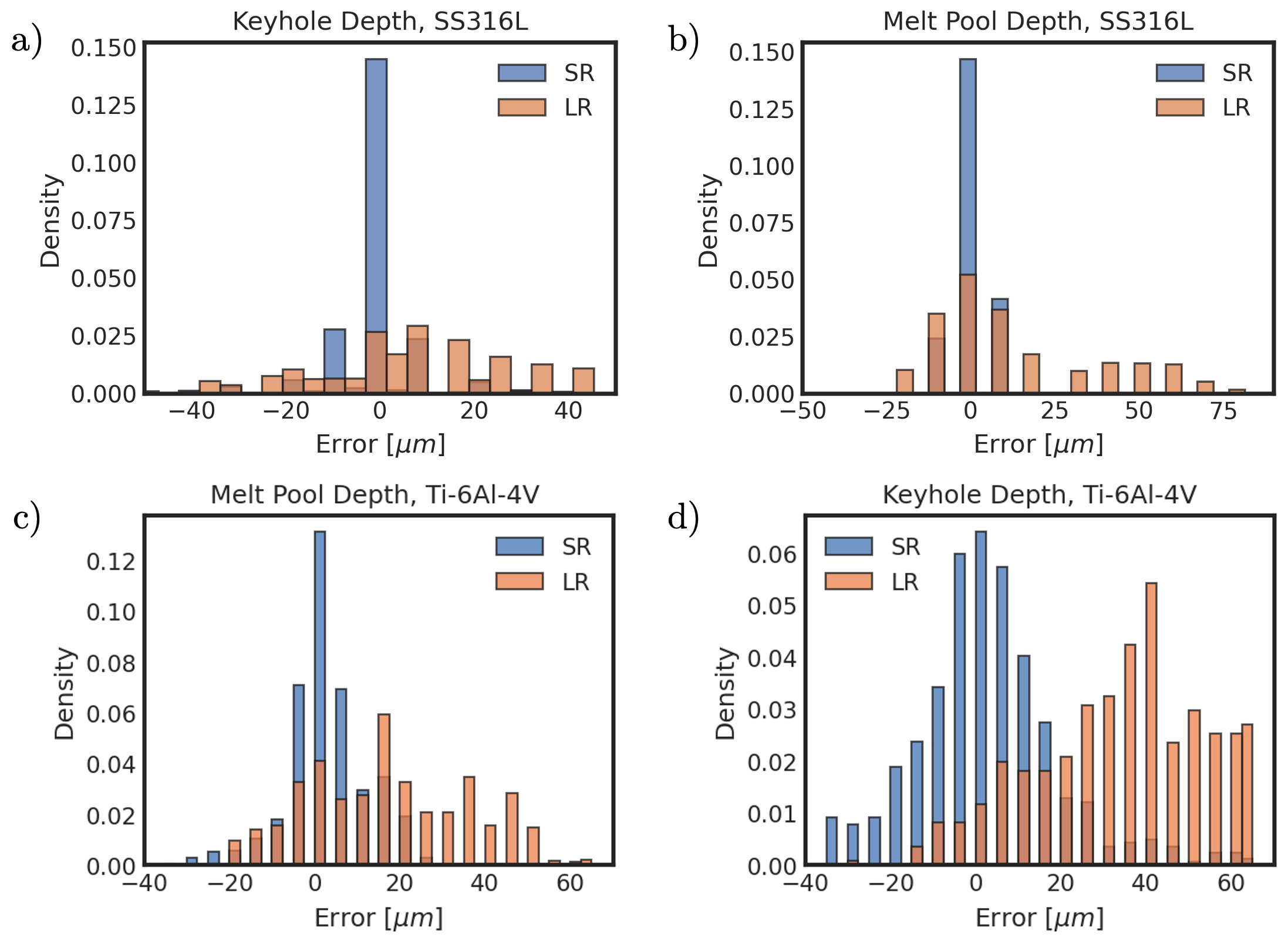}
   \caption{The probability density distribution of error of the different prediction tasks represented visually. LR (low-resolution) is the error of comparing the bicubic interpolation of the low-fidelity to the model to the high-fidelity ground truth, and SR (super-resolution) is the error of comparing the diffusion model upscaling output to the ground truth. a) The aggregate error of the melt pool depth for the model trained on the SS316L $2\times$ upscaling task. b) The aggregate error of the keyhole depth for the model trained on the SS316L $2\times$ upscaling task. c) The aggregate error of the melt pool depth for the model trained on the Ti-6Al-4V $4\times$ upscaling task. d) The aggregate error of the keyhole depth for the model trained on the Ti-6Al-4V $4\times$ upscaling task.}
    \label{fig:diffusion_error_histogram}
\end{figure}

Next, the reconstruction output of the 2$\times$ upscaling tasks is quantitatively evaluated based on the error in the reconstructed melt pool dimensions, the reconstructed keyhole dimensions, and temperature field error. The bicubic interpolation error is compared to the diffusion model error visually for both the 4$\times$ and 2$\times$ upscaling tasks in Figure \ref{fig:diffusion_error_histogram}.  Due to the stochastic nature of the prediction process, the diffusion MAE is slightly higher than the CNN MAE, as the CNN is directly trained to minimize the MAE over the dataset while sacrificing image quality and realism. However, the diffusion model employs stochastic sampling to generate new images, which enables it to resolve more realistic, feasible keyhole structures. Therefore, we again benchmark the ability of the diffusion model to recreate the melt pool and keyhole variability, and display the results in Figure \ref{fig:ss316ldiffusionfreqamp}. From this figure, we observe that the sampling of the high fidelity data distribution yields keyhole depth values that fall within the range of the ground truth keyhole oscillation. To show this quantitatively, we plot the predicted and simulated keyhole oscillation amplitude in Figure \ref{fig:ss316ldiffusionfreqamp}c, achieving a Pearson $R^2$ coefficient of 0.74 for the diffusion output, much larger than the  $R^2$ of 0.06 achieved with the CNN model.

\begin{table}[]
\caption{Metrics evaluating the performance of the  Diffusion model on the 2$\times$ upscaling task.}
\centering
\begin{tabular}{@{}lllll@{}}

\toprule
         Upsampling Method            & MAE [K] & MP-MAE [$\mu m$] & VC-MAE [$\mu m$] \\ \midrule
Bicubic Upsampling   & 79.2        &    19.5   &   32.7   \\
CNN Upsampling     &  31.7      & 3.20   &    7.79   \\
Diffusion Upsampling  &39.3  &   2.91  & 8.89    \\ \bottomrule
\label{table:ss316l_metrics}
\end{tabular}
\end{table}
\FloatBarrier
\subsection{Computational Details}

\begin{figure}[htbp!]
    \centering
    \includegraphics[width=0.6\textwidth]{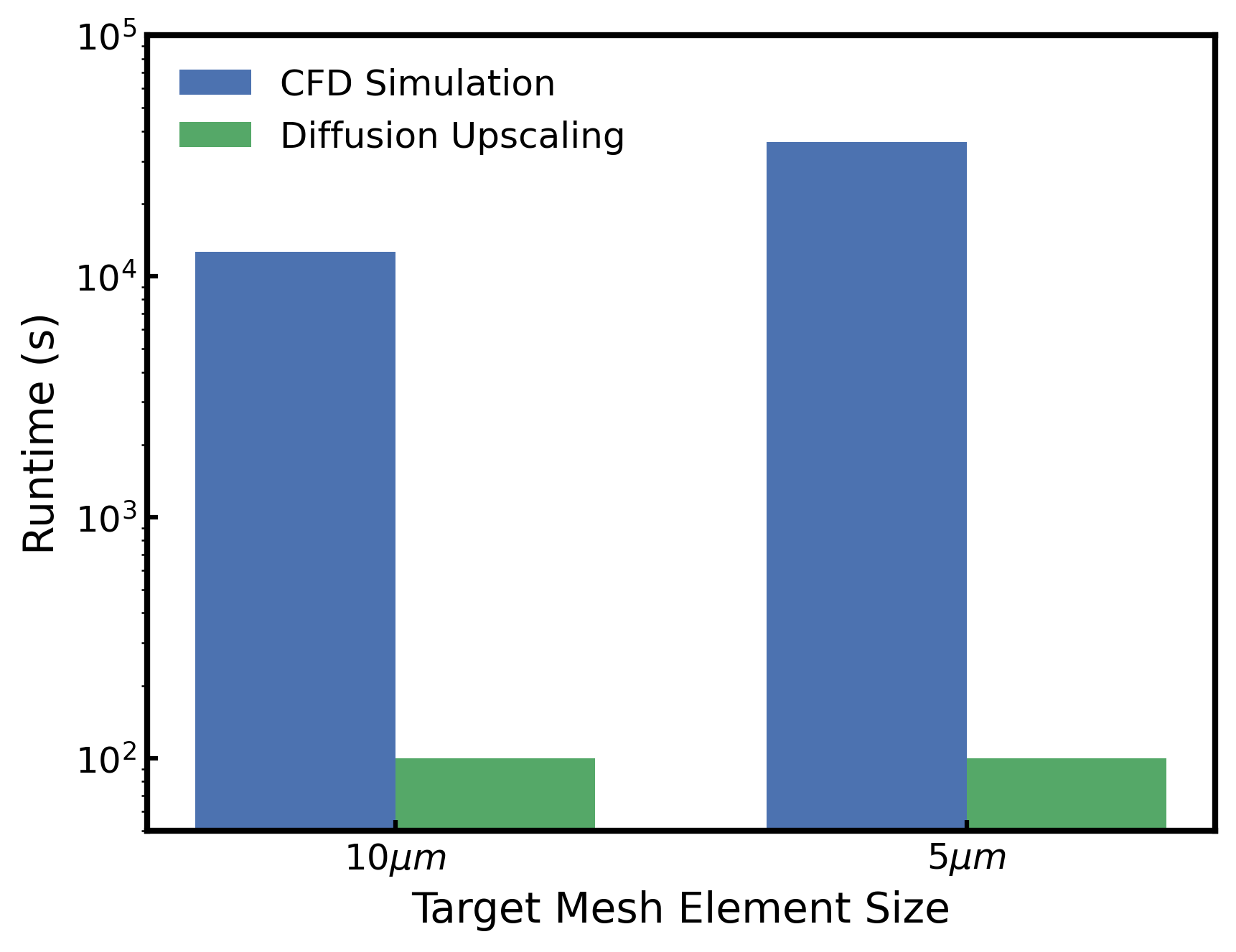}
   \caption{A time comparison of running a full-field CFD simulation at the specified mesh element size in FLOW-3D, compared to the time required to run a simulation at a $20 \mu m$ mesh and perform diffusion upscaling on each timestep of the 2-D cross-section of the simulation.}
    \label{fig:time_comparison_seconds}
\end{figure}

The FLOW-3D simulations were run using a workstation with a 16 core Intel i7-10700F 2.90 GHz processor.
The training and inference processes are carried out on an NVIDIA GeForce RTX-2080 GPU with a 11 GiB memory capacity. Using the aforementioned computational configuration, direct simulations using FLOW-3D at a 5 $\mu m$ mesh element resolution require three hours of run time per simulation on average. A single epoch (165 iterations of a batch of size 128) is trained in 1.67 minutes for the CNN encoder model, and 1.51 minutes for the diffusion U-Net model. Therefore, the overall training process is completed in 32 hours. The inference process, accelerated by the modified sampler, is completed in 0.41 seconds for a single sample. A comparison of inference time against the total simulation time for the resolutions studied in this work is presented in Figure \ref{fig:time_comparison_seconds}..

% \begin{figure}[h]
%     \centering
%     \includegraphics[width=\textwidth]{figures/results_multifield.png}
%     \caption{}
%     \label{fig:multifield_diffusion_results}
% \end{figure}
% \section{Conclusion}
% \label{sec:sample2}
\FloatBarrier
\section{Conclusion}
\label{sec:conclusion}
In this work, a method for deep-learning based spatial upsampling of melt pool temperature distributions produced during laser powder bed fusion is presented. To do so, a generative model is trained to capture the distribution of potential high-fidelity outputs possible given a sample created with a low-fidelity model of the melt pool process. We generate these low-fidelity models of the melt pool by running low-fidelity simulations of the melt pool behavior, where complex high-frequency phenomena is unlikely to be fully resolved. To upsample the data while accounting for the large scale morphology differences between the low-fidelity input and the high-fidelity target fields, we first employ a CNN encoder that predicts the mean of the distribution of the potential high-fidelity output fields. The embedded features learned by this CNN are then taken as the conditioning input for the diffusion model. This model stabilizes training by reducing the distance from the empirical data distribution of the high fidelity target fields from the conditioning used as input. 

The performance of this model is demonstrated on two tasks, a $2\times$ upscaling from SS316L simulations ran at $20\mu m$ and 10$\mu m$ resolutions respectively, and a $4\times$ upscaling from Ti-6Al-4V simulations ran at 20$\mu m$ and 5$\mu m$ respectively. Due to the fact that the diffusion model is a generative model that parameterizes the empirical high-fidelity data distribution, the individual samples produced by a given trained model are stochastic. Thus, conventional metrics, such as the MAE, are misleading in this application. However, the MAE is reduced from non-deep learning based approaches, such as bicubic upscaling. To capture the performance of the model in a more principled manner, the variability of the keyhole behavior and the dimensions of the melt pool are examined as predictors for the onset of porosity. In these cases, the proposed diffusion model is better able to capture the intrinsic variability of these parameters, producing superior image quality. The upscaling process enables implicit information about varying process and operating conditions to be embedded in the low-fidelity simulation physics, as opposed to being explicitly provided to the model as manually constructed feature information. Therefore, this framework accelerates studies of the process-structure-property relationship, by enabling faster parameter sweeps and extraction of high-fidelity information from lightweight computational simulations. Alternatively, extensions of this model to three-dimensional training data can enhance multi-scale modeling approaches, integrating directly into large-scale coarse simulations to provide information on the small-scale effects that would otherwise be neglected. 

% Generally, the behavior of a case with an arbitrary process parameter conditions that lie within the training process space boundaries can be reasonably approximated, avoiding the computational expense of high-fidelity simulations. By modeling the statistical relationship between the low-fidelity model response and the probable corresponding high-fidelity model responses after training, the model can approximate the high-fidelity behavior of the process and provide a confidence interval describing the range of feasible model outputs.

%A method for upscaling melt pool temperature fields from low-resolution cross-sections is presented.
%The melt pool temperature dynamics and key morphological details are preserved with the diffusion model.

\section*{Acknowledgments}
Research was sponsored by the Army Research Laboratory, USA and was accomplished under Cooperative Agreement Number W911NF-20-2-0175. The views and conclusions contained in this document are those of the authors and should not be interpreted as representing the official policies, either expressed or implied, of the Army Research Laboratory or the U.S. Government. The U.S. Government is authorized to reproduce and distribute reprints for Government purposes notwithstanding any copyright notation herein.

\pagebreak
 \bibliographystyle{elsarticle-num} 
 \bibliography{cas-refs}
 \pagebreak
\appendix

\section{Additional Dataset Details}
\label{sec:sample:appendix}
% \subsection{Dataset Details}
\subsubsection*{Processing Parameters}

\begin{figure}[htbp!]
    \centering
    
    \includegraphics[width=\textwidth]{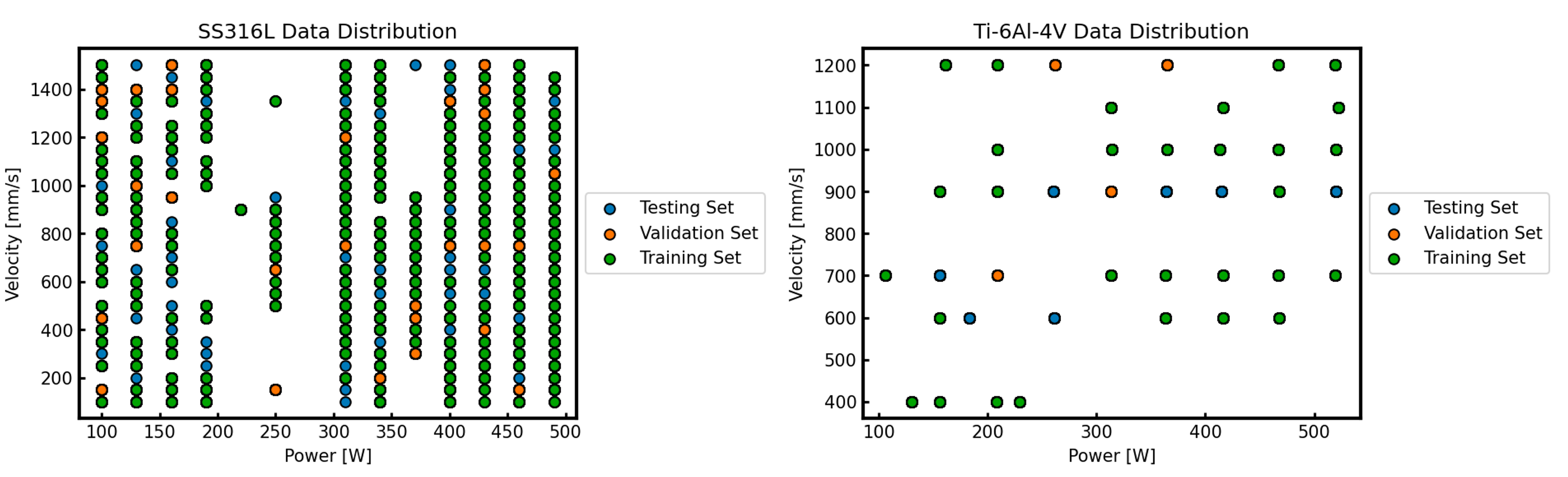}
    \caption{(Left) The distribution of power and velocity combinations used to train and evaluate the SS316L 2$\times$ upscaling task. (Right) The distribution of power and velocity combinations used to train and evaluate the Ti-6Al-4V 4$\times$ upscaling task. }
    \label{fig:datadescription}
\end{figure}

\begin{figure}[htbp!]
    \centering
    
    \includegraphics[width=0.65\textwidth]{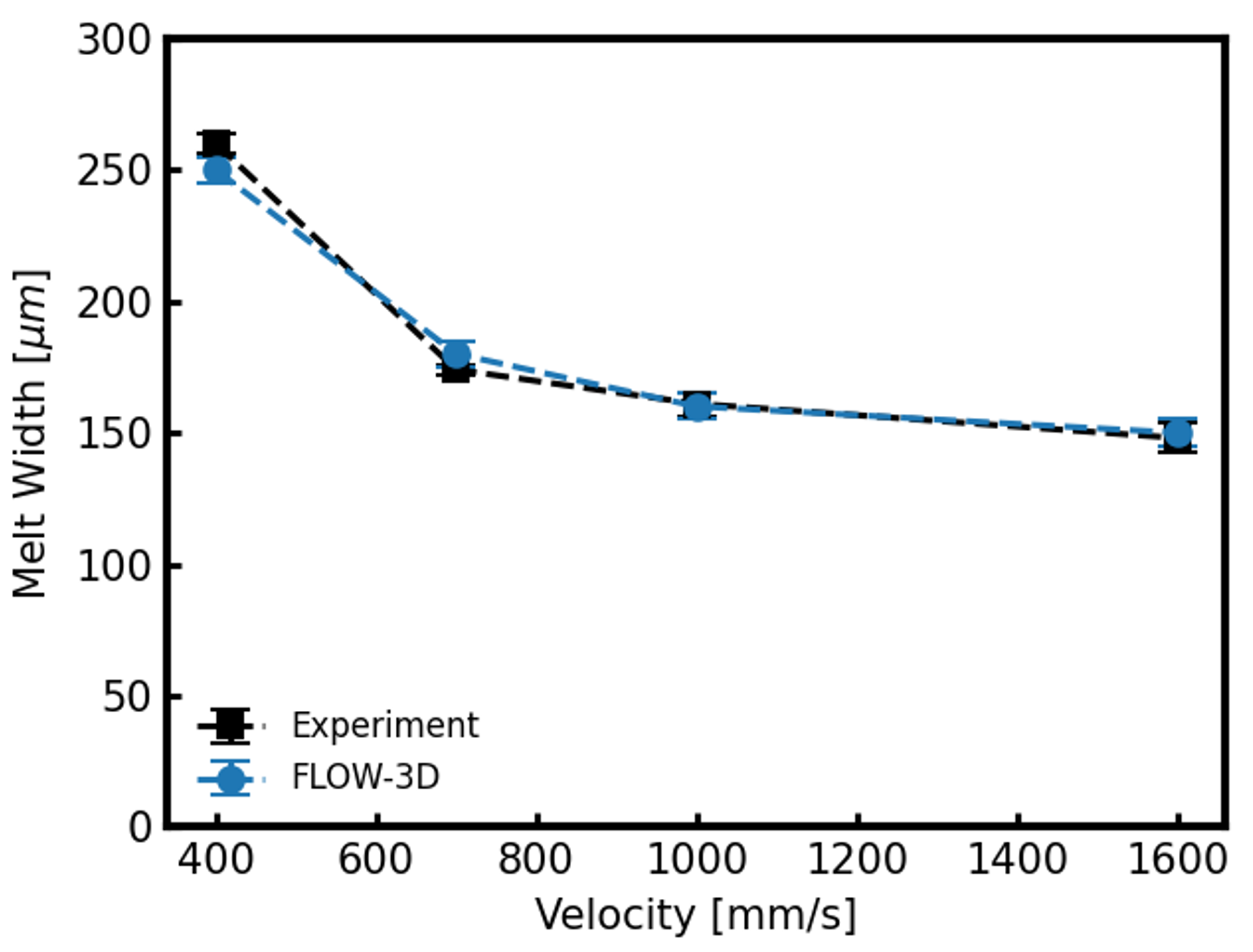}
    \caption{The width of single bead tracks at varying velocities produced with the SS316L parameters specified in Table \ref{table:ss316lmaterialparam}. These are compared to experimental width values found after printing on a TRUMPF TruPrint 3000 Laser Powder Bed Fusion machine. The laser power is fixed to 300 W and the laser diameter is fixed to 100 $\mu m$ for all displayed data points. Further information about the configuration of the L-PBF experiments is available in Ref. \cite{myers2023high}. }
    \label{fig:validation_mp_width}
\end{figure}

\FloatBarrier

\subsubsection*{Material Properties}

\begin{table}[ht!]
\centering

\label{tableti64:materialparam}
\caption{Material Parameters used to simulate the Ti-6Al-4V melting process}
\begin{tabular}{@{}lllll@{}}
\toprule
Parameter                        & Value                                            & Units                   &  &  \\ \midrule
Density, $\rho$, 298 K              & 4420                                             & kg/m$^3$ &  &  \\
Density, $\rho$, 1923 K             & 3920                                             & kg/m$^3$ &  &  \\
Specific Heat, $C_v$, 298 K         & 546                                              & J/kg/K                  &  &  \\
Specific Heat, $C_v$, 1923 K        & 831                                              & J/kg/K                  &  &  \\
Vapor Specific Heat, $C_{v, vapor}$ & 600                                              & J/kg/K                  &  &  \\
Thermal Conductivity, $k$, 298 K    & 7                                                & W/m/K                   &  &  \\
Thermal Conductivity, $k$, 1923 K  & 33.4                                             & W/m/K                   &  &  \\
Viscosity, $\eta$                        & 0.00325                                          & kg/m/s                  &  &  \\
Surface Tension, $\sigma$          &1.882  &       $kg/s^2$            &      &  \\
Liquidus Temperature, $T_L$       & 1923                                             & K                       &  &  \\
Solidus Temperature, $T_S$        & 1873                                             & K                       &  &  \\
Fresnel Coefficient $\epsilon$        & 0.25 &         -           &  &  \\
Accommodation Coefficient, $a$         &0.005  &-                    &  &  \\
Latent Heat of Fusion, $\Delta H_f$            & 2.86 $\times 10^5$ & J/kg                    &  &  \\

Latent heat of vaporization,  $\Delta H_v$     & 6.00 $\times 10^4$ & J/kg                    &  &  \\ \bottomrule
\end{tabular}
\end{table}

\begin{table}[ht!]
\centering
\caption{Material Parameters used to simulate the SS316L melting process}
\begin{tabular}{@{}lllll@{}}
\toprule
Parameter                        & Value                                            & Units                   &  &  \\ \midrule
Density, $\rho$, 298 K              & 7950                                             & kg/m$^3$ &  &  \\
Density, $\rho$, 1923 K             & 6765                                            & kg/m$^3$ &  &  \\
Specific Heat, $C_v$, 298 K         & 470                                              & J/kg/K                  &  &  \\
Specific Heat, $C_v$, 1923 K        & 1873                                              & J/kg/K                  &  &  \\
Vapor Specific Heat, $C_{v, vapor}$ & 449                                              & J/kg/K                  &  &  \\
Thermal Conductivity, $k$, 298 K    & 13.4                                                & W/m/K                   &  &  \\
Thermal Conductivity, $k$, 1923 K  & 30.5                                            & W/m/K                   &  &  \\
Viscosity, $\eta$                        & 0.008                                         & kg/m/s                  &  &  \\
Surface Tension, $\sigma$          & 1.882 &          $kg/s^2$          &  &  \\
Liquidus Temperature, $T_L$       & 1723                                             & K                       &  &  \\
Solidus Temperature, $T_S$        & 1658                                             & K                       &  &  \\

Fresnel Coefficient, $\epsilon$         &0.2  &         -           &  &  \\

Accommodation Coefficient, $a$          &0.15  & -                    &  &  \\

Latent Heat of Fusion, $\Delta H_f$            & 2.6 $\times 10^5$ & J/kg                    &  &  \\
Latent Heat of Vaporization, $\Delta H_v$    & 7.45 $\times 10^6$ & J/kg                    &  &  \\ \bottomrule
\end{tabular}
\label{table:ss316lmaterialparam}

\end{table}
\FloatBarrier
\section{Additional Diffusion Model Details}
\subsubsection*{Model Architecture Structure}

Below, we evaluate the impact of different architectural choices for conditioning the diffusion model based on the corresponding performance on the upscaling task. Conventional computer vision implementations of superresolution using diffusion models concatenate the bicubic interpolation of the low fidelity input to the sampled noise vector as the input to the diffusion model. We evaluate the influence of using the up-scaled bicubic interpolation of the low fidelity as the conditioning step, compared to the effect of using the output of the CNN model as the conditioning input. From Table \ref{table:meltpoolgeometrymetrics}, we can observe that the encoder model is necessary in order to extract the relevant information from the low fidelity input for a mapping to the high fidelity space. The use of a convolution layer and addition of the noise vector to the conditioning input, adapted from Ref. \cite{li2022srdiff} as opposed to concatenation also yields an increase in performance.
\begin{table}[htbp!]
\centering

\caption{Melt Pool Geometry Metrics based on the diffusion model conditioning format}
\begin{tabular}{@{}llll@{}}

\toprule
  Conditioning Method                                    & MP-MAE & VC-MAE &  \\ \midrule
Encoder, Addition                     & 3.58   & 12.7   &  \\
Encoder, Concatenation                & 3.72   & 12.9   &  \\
Bicubic Interpolation, Addition       & 7.71   & 15.2   &  \\
Bicubic Interpolation, Concatentation & 52.0   & 28.3   &  \\ \bottomrule
\end{tabular}
\label{table:meltpoolgeometrymetrics}
\end{table}
\pagebreak
\FloatBarrier
\subsubsection*{Diffusion Timesteps}
\begin{figure}[htbp!]
    \centering
    \includegraphics[width=\textwidth]{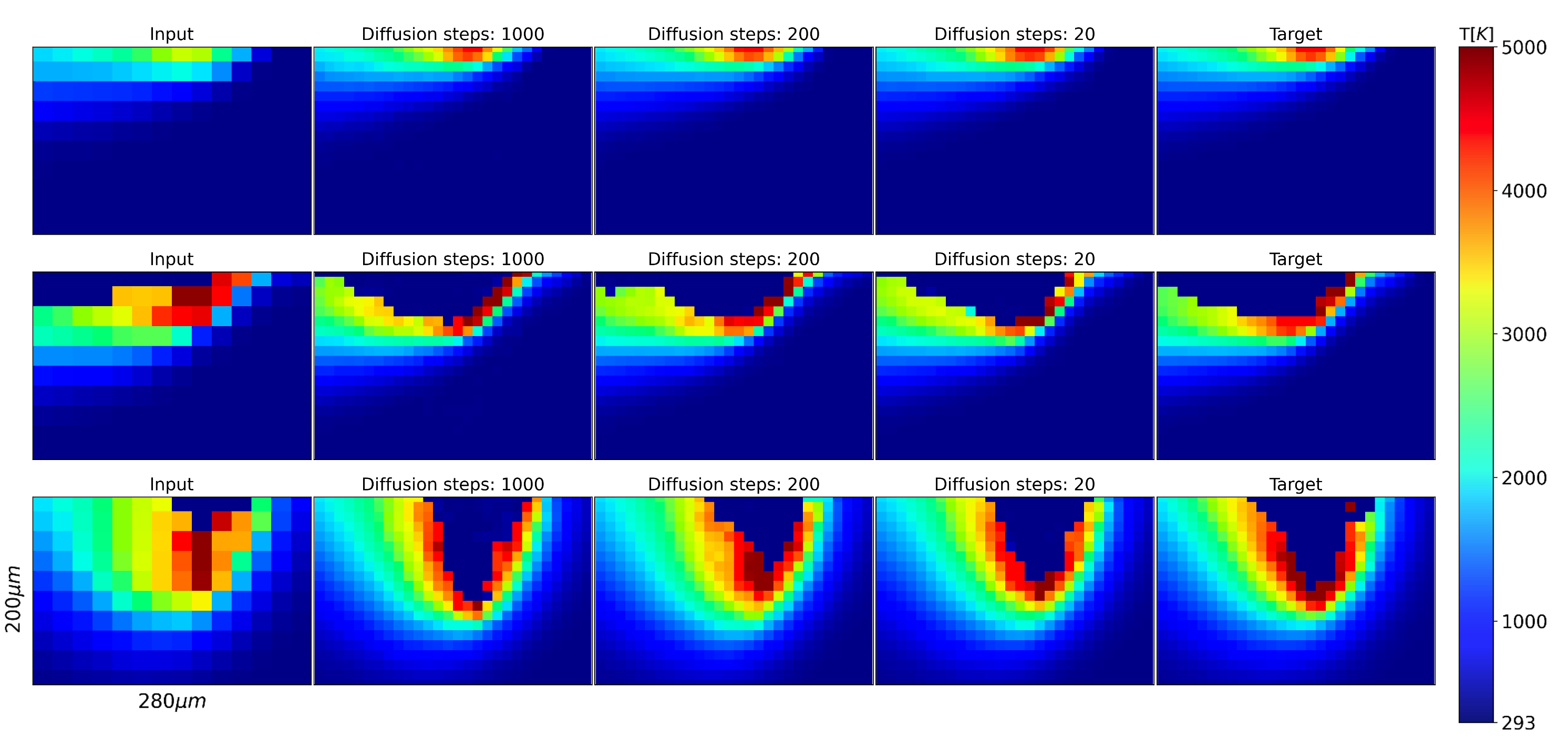}
  \caption{An ablation study on the number of timesteps used during the denoising sampling process. The image quality is effectively independent of the number of timesteps used for the range studied in this figure. This allows us to use 20 timesteps as opposed to 1000 timesteps for denoising, accelerating the sampling process by 50$\times$.}
    \label{fig:diffusion_step_ablation}
\end{figure}
\FloatBarrier
\subsubsection*{Diffusion Uncertainty}
\begin{figure}[htbp!]
    \centering
    \includegraphics[width=\textwidth]{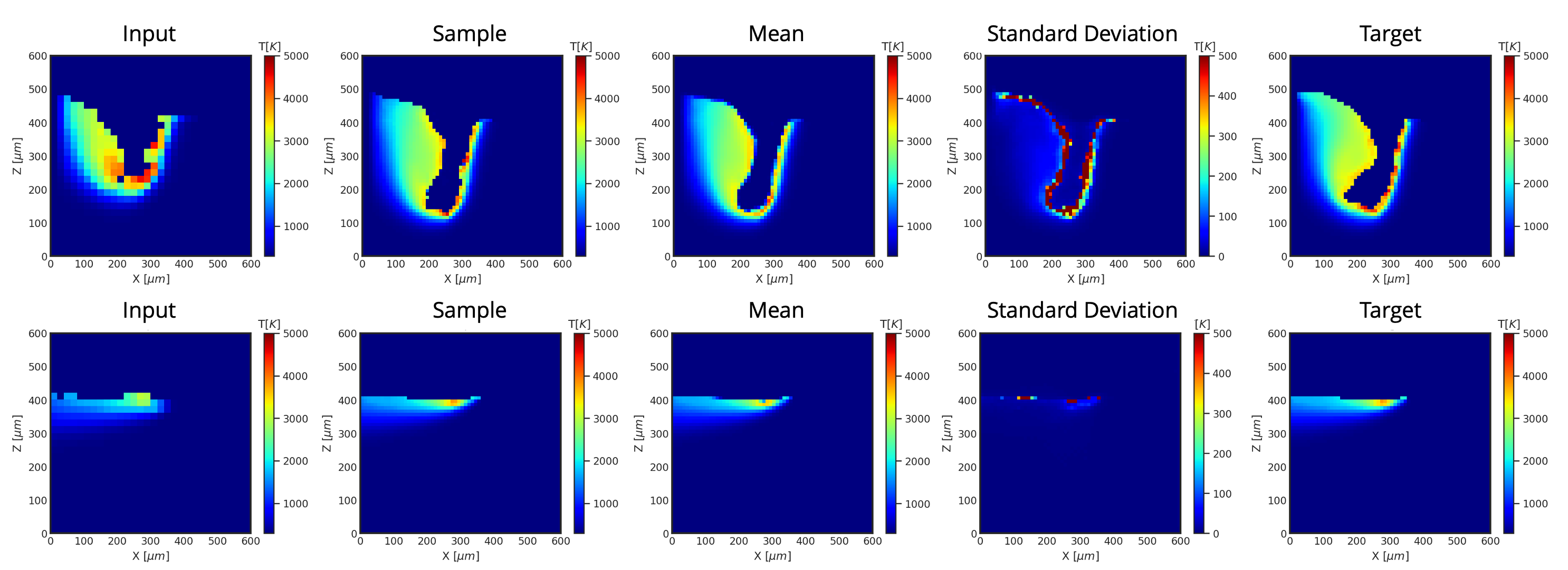}
  \caption{A visualization of the distribution learned by the diffusion model, demonstrated on two samples from the SS316L dataset. High frequency areas of the melt pool, such as the oscillating free surface are represented by a high variance in the diffusion output, while areas of the melt pool that change on longer timescales have a much smaller variance. The diffusion model samples from this empirical distribution at inference time. }
    \label{fig:diffusion_std_mean}
\end{figure}
\FloatBarrier






% \end{thebibliography}
\end{document}